\documentclass{article} %
\usepackage{iclr2024_conference,times}

\newcommand{\reb}[1]{{#1}}

\newif\ifCommentedIII

\CommentedIIIfalse

\ifCommentedIII
\newcommand{\ptdyIII}[1]{\textcolor{teal}{#1}} 
\newcommand{\fdyIII}[1] 
{{\color{teal}\footnote{\textcolor{teal}{DY: #1}}}}
\newcommand{\fbtkIII}[1]{{\color{red}\footnote{\textcolor{red}{TK: #1}}}}
\newcommand{\fgkIII}[1]{{\color{orange}\footnote{\textcolor{orange}{GK: #1}}}}
\newcommand{\ptmdIII}[1]{\textcolor{blue}{#1}} 
\newcommand{\fmdIII}[1]{{\color{blue}\footnote{\textcolor{blue}{MD: #1}}}}
\newcommand{\fjrIII}[1]{{\color{yellow}\footnote{\textcolor{yellow}{JR: #1}}}}
\else
\newcommand{\ptdyIII}[1]{#1}
\newcommand{\fdyIII}[1]{}

\newcommand{\fbtkIII}[1]{}

\newcommand{\fgkIII}[1]{}
\newcommand{\ptmdIII}[1]{#1}
\newcommand{\fmdIII}[1]{}

\newcommand{\fjrIII}[1]{}
\fi

\newif\ifCommentedII

\ifCommentedII
\newcommand{\ptdyII}[1]{\textcolor{teal}{#1}} 
\newcommand{\fdyII}[1] 
{{\color{teal}\footnote{\textcolor{teal}{DY: #1}}}}
\newcommand{\fbtkII}[1]{{\color{red}\footnote{\textcolor{red}{TK: #1}}}}
\newcommand{\fgkII}[1]{{\color{orange}\footnote{\textcolor{orange}{GK: #1}}}}
\newcommand{\ptmdII}[1]{\textcolor{blue}{#1}} 
\newcommand{\fmdII}[1]{{\color{blue}\footnote{\textcolor{blue}{MD: #1}}}}
\newcommand{\fjrII}[1]{{\color{yellow}\footnote{\textcolor{yellow}{JR: #1}}}}
\else
\newcommand{\ptdyII}[1]{#1}
\newcommand{\fdyII}[1]{}

\newcommand{\fbtkII}[1]{}

\newcommand{\fgkII}[1]{}
\newcommand{\ptmdII}[1]{#1}
\newcommand{\fmdII}[1]{}

\newcommand{\fjrII}[1]{}
\fi

\newif\ifCommentedI

\CommentedIfalse   

\ifCommentedI
\newcommand{\ptdyI}[1]{\textcolor{teal}{#1}} 
\newcommand{\fdyI}[1] 
{{\color{teal}\footnote{\textcolor{teal}{DY: #1}}}}
\newcommand{\fbtkI}[1]{{\color{red}\footnote{\textcolor{red}{TK: #1}}}}
\newcommand{\fgkI}[1]{{\color{orange}\footnote{\textcolor{orange}{GK: #1}}}}
\newcommand{\ptmdI}[1]{\textcolor{blue}{#1}} 
\newcommand{\fmdI}[1]{{\color{blue}\footnote{\textcolor{blue}{MD: #1}}}}
\newcommand{\fjrI}[1]{{\color{yellow}\footnote{\textcolor{yellow}{JR: #1}}}}
\else
\newcommand{\ptdyI}[1]{#1}
\newcommand{\fdyI}[1]{}

\newcommand{\fbtkI}[1]{}

\newcommand{\fgkI}[1]{}
\newcommand{\ptmdI}[1]{#1}
\newcommand{\fmdI}[1]{}

\newcommand{\fjrI}[1]{}
\fi

\usepackage{amsmath,amsfonts,bm}

\def\eqref#1{equation~\ref{#1}}

\def\1{\bm{1}}

\DeclareMathAlphabet{\mathsfit}{\encodingdefault}{\sfdefault}{m}{sl}
\SetMathAlphabet{\mathsfit}{bold}{\encodingdefault}{\sfdefault}{bx}{n}

\usepackage{hyperref}
\usepackage{url}

\usepackage{multirow}
\usepackage{makecell}
\usepackage{booktabs}
\usepackage{amsmath,amsfonts,amssymb}
\usepackage{graphicx}
\usepackage{inconsolata} %
\usepackage{array}
\newcolumntype{P}[1]{>{\centering\arraybackslash}p{#1}}
\newcolumntype{M}[1]{>{\centering\arraybackslash}m{#1}}
\usepackage{comment}
\usepackage{wrapfig}

\title{Compositional preference models \\for aligning LMs}

\author{Dongyoung Go \\
Naver Corp\\
Yonsei University\\
\texttt{dongyoung.go@navercorp.com} \\
\And
Tomasz Korbak \\
University of Sussex \\
\texttt{tomasz.korbak@gmail.com}\hspace{22px} \\
\AND
Germán Kruszewski, Jos Rozen \\
Naver Labs Europe \\
\texttt{\{german.kruszewski,jos.rozen\}@naverlabs.com}
\And
Marc Dymetman \\
Independent Researcher \\
\texttt{marc.dymetman@gmail.com}
}

\iclrfinalcopy %
\begin{document}

\maketitle

\begin{abstract}
As language models (LMs) become more capable, it is increasingly important to align them with human preferences. However, the dominant paradigm for training Preference Models (PMs) for that purpose suffers from fundamental limitations, such as lack of transparency and scalability, along with susceptibility to overfitting the preference dataset. We propose Compositional Preference Models (CPMs), a novel PM framework that decomposes one global preference assessment into several interpretable features, obtains scalar scores for these features from a prompted LM, and aggregates these scores using a logistic regression classifier. 
\ptdyIII{Through these simple steps,}
CPMs allow to control which properties of the preference data are used to train the preference model and to build it based on features that are believed to underlie the human preference judgement.
Our experiments show that CPMs not only improve generalization and are more robust to overoptimization than standard PMs, but also that best-of-$n$ samples obtained using CPMs tend to be preferred over samples obtained using conventional PMs.
Overall, our approach demonstrates the benefits of endowing PMs with priors about which features determine human preferences while relying on LM capabilities to extract those features in a scalable and robust way.%
\end{abstract}

\begin{wrapfigure}{t}{0.51\textwidth}
    \centering
    \vspace{-0.48cm}
    \includegraphics[width=1\linewidth]{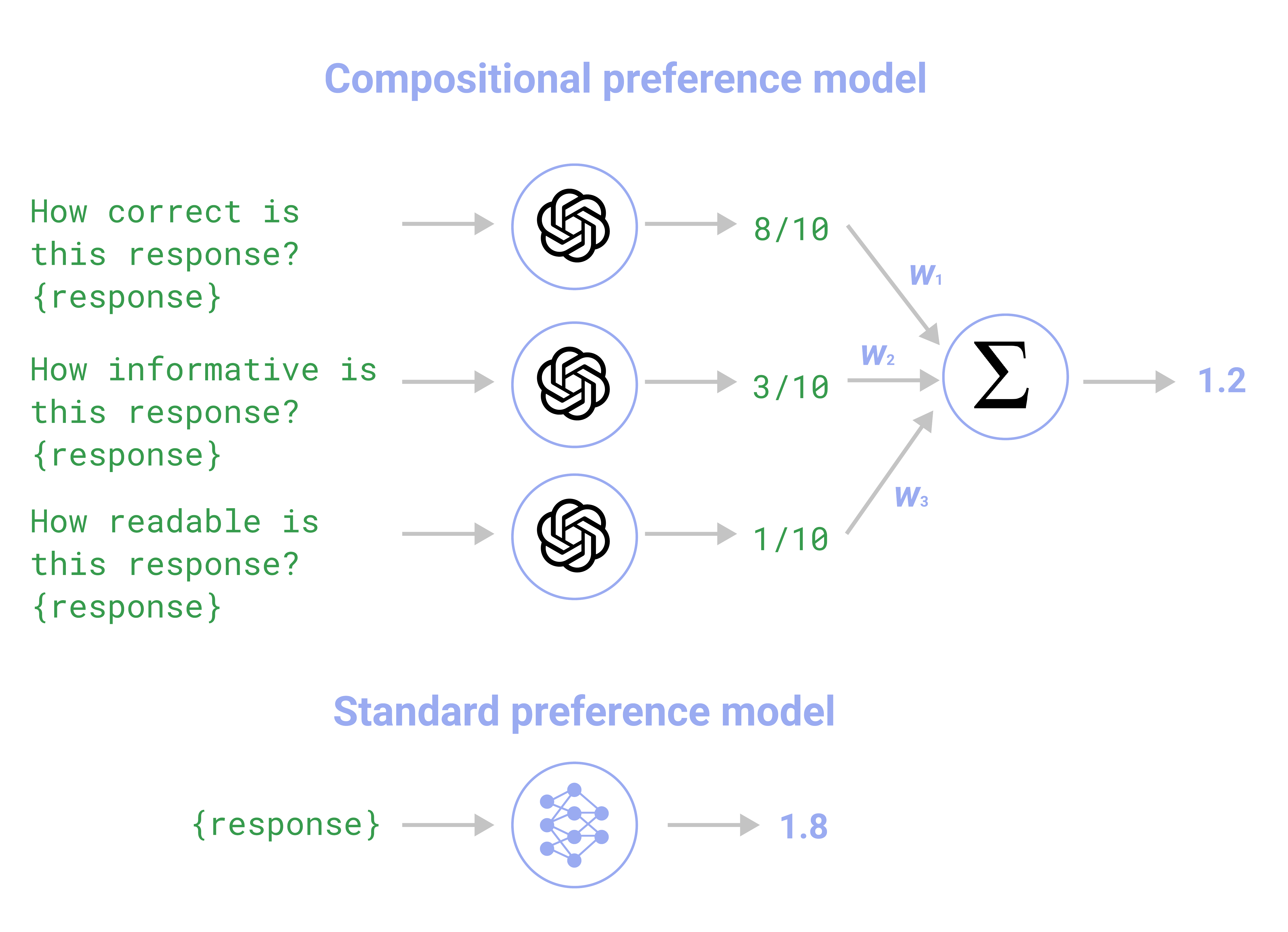} 
    \caption{Compositional preference models score different features of LM responses separately and output a preference score as a linear combination of feature values.
    }
    \label{fig:main}
\end{wrapfigure}

\section{Introduction}

As the capabilities of language models (LMs) continue to advance, there is a growing need for safe and interpretable \ptmdI{models}.
The dominant approach to aligning LMs with human preferences,  reinforcement learning from human feedback \citep[RLHF;][]{ouyang2022training, bai2022training, openai2023gpt4}, consists in training a preference model (PM) to predict human preference judgments and then finetuning an LM to maximize \ptmdI{the} reward given by the PM.
However, the current PM methodology exhibits certain limitations. 
First, \ptmdII{it is} susceptible to overfitting the preference dataset. 
The PM can misrepresent human preferences by fitting to spurious correlations in its training data \cite{gao2023scaling}. Heavily optimizing an LM against a PM incentivises the LM to exploit those flaws. This effect is known as reward hacking or Goodhart’s law \citep{goodhart1984problems}. 
One way of addressing reward hacking is \ptmdII{to impose} certain inductive biases on the PM or limiting its capacity. Second, PMs are often difficult to interpret and 
\ptdyI{to oversee}
\fgkI{I didn't get the feedback part of this sentence\ptdyI{DY: basically it was for introducing scalable oversight}}. 
They project preferences onto a single scalar feature, making it difficult to know what factors are influencing their decisions. This is especially problematic for complex preferences, such as helpfulness or harmlessness, which often encompass a multidimensional combination of attributes \citep{bai2022training, glaese2022improving, touvron2023llama}.
\ptdyII{Further, a}s LM capabilities \ptmdII{improve}, it will be increasingly harder for unassisted humans to provide feedback on LM’s responses \citep{pandey2022modeling,bowman2022}.
One way of addressing this problem is \ptmdII{to use} another LM to decompose those responses into simpler pieces that can be evaluated either by a human or \ptmdI{an} LM.

In this paper, we propose \ptmdI{the} Compositional Preference Model (CPM), a novel framework for learning a PM that is robust to preference model overoptimization and allows for more transparent and interpretable supervision of complex behavior. \ptmdI{A} CPM decomposes one global preference assessment into a series of simpler questions which correspond to human-interpretable features. 
Then, a prompted LM (e.g. 
\ptmdII{GPT-3.5}) is asked to assign a numerical value to each feature. Finally, the feature scores are combined into \ptmdI{a} scalar preference score using a trained logistic regression classifier.

CPM\ptmdII{s} have several advantages over standard PMs. First, \ptmdII{they} are more robust to overfitting and reward hacking. The pre-selected features on which CPM\ptmdI{s} operate provide a useful inductive bias that bootstraps learning human preferences.
This, in turn, limits \ptmdI{their} vulnerability to reward hacking, as the parameter space of a PM is spanned by features selected to be meaningful and robust. Second, CPMs allow for the modular and human-interpretable supervision of complex behavior. \ptmdI{They} effectively decompose a hard question (e.g. ``is this text preferable?”) into a series of easier questions (e.g. ``is this text \ptdyIII{easy to read?”,}
``is this text informative?”) that are easier to evaluate for an LM and easier to inspect for a human overseer. 
This is a simple instance of a divide-and-conquer supervision approach \citep{cormen2022introduction},  which recursively breaks down a problem until it is easily solvable and then combines the solutions \citep{irving2018ai,leike2018scalable,christiano2018supervising}.

In our experiments, we show that 
CPMs generalize better and \ptmdI{that} using them results in less preference model overoptimization. 
Additionally, CPMs exhibit superior performance in capturing the underlying 
\ptdyII{human preferences.}
In \ptmdI{an} auto-evaluation experiment with Claude \citep{claude} as an approximation of human evaluators \citep{vicuna2023,mukherjee2023orca,liu2023gpteval,he2023annollm},
best-of-$n$ samples obtained using CPMs  are consistently preferred over samples obtained using conventional PMs.\footnote{Code accompanying the paper is available at \href{https://github.com/dongyoung-go/CPM}{https://github.com/dongyoung-go/CPM}}

Overall, the contributions of the paper include:
\vspace{-5px}
\begin{enumerate}
\itemsep0em 
    \item Introducing CPM, a novel framework for learning PMs that is more robust to overoptimization and allows for more transparent 
    supervision, by decomposing \ptmdI{the} preference problem into \ptmdI{a} series of intuitive features linked to human preferences, and employing \ptmdI{an} LLM as a feature score extractor (Sec.~\ref{sec:method}).
    \item Investigating the performance of CPM\ptmdII{s} on a diverse array \ptmdI{of dimensions},  including model robustness (Sec.~\ref{sec:model_robustness}), generalization (Sec.~\ref{sec:referece_comparison}), robustness to overoptimization (Sec.~\ref{sec:overoptimization}), and effectiveness for preference alignment (Sec.~\ref{sec:quality_eval}).
    \item Enabling \ptmdI{an} intuitive explanation of model optimization and generated responses (Sec.~\ref{sec:interpretability}).
\end{enumerate}

\section{Background}\label{sec:preliminaries}

Let us have a dataset of comparisons 
$\mathit{\mathcal{D}}=\{x^{i},y_{1}^{i},y_{2}^{i}\}_{i=1}^{N}$, 
where $x$ \ptmdII{is} an input query and $y_1$ and $y_2$ are two possible responses to $x$, with $y_1$ the preferred response.
The dominant approach to aligning language models, RLHF 
\citep{christiano2017deep,ziegler2019fine,ouyang2022training, bai2022training}\footnote{{CPMs can also be used with other 
\ptdyII{alignment}
training methods both during pretraining \citep{korbak23_pretraining} and finetuning \citep{rafailov2023direct,pmlr-v202-go23a}.}},
\ptmdII{involves}
training a parametrized PM $R(y|x) = R_\theta(y|x)$ by defining a probability distribution
\vspace{-0.45px}
\begin{equation}\label{eq:standard-PM}
p_\theta(y_{1}>y_{2}|x) \doteq \sigma(R_\theta(y_{1}|x)-R_\theta(y_{2}|x)) = (1+\exp(R_\theta(y_{2}|x)-R_\theta(y_{1}|x))^{-1}
\end{equation}
and estimating $\theta$ by maximizing the likelihood of $p_\theta$ over $\mathcal{D}$. 
Typically $R_\theta$ is obtained by adding a scalar head on top of a base language model and fine-tuning the resulting model. Since \ptmdII{$p_\theta$ is invariant to addition of a constant to $R_\theta$}, it is standard to shift the \ptmdII{$R$} scores such that $E_{(x,y)\sim D}[R(y|x)]=0$.

\section{Method}\label{sec:method}

The Compositional Preference Model (CPM) \ptmdII{is} a multi-step approach for decomposing preference learning into individual components. 
We first decompose %
preference judgements
into a set of \ptmdII{$C$} distinct features, each designed to evaluate a specific aspect of the response $y$ (relative to context $x$). 
\ptmdII{Then we} use a prompted LM to assign to a pair $(x, y)$ a scalar score 
for each individual
feature 
\ptdyI{$c=1,\ldots,C$}.
\ptmdII{Finally, we} employ a logistic regression classifier 
to combine these features into a global scalar score that best predicts the human preference judgements. This approach enables us to construct a coherent description of the characteristics 
that 
underlie
these judgements.

\subsection{Feature extraction using a language model}\label{sec:feature_extraction}
 
For each feature $c$, we consider an individual preference model $R_{c}$ that maps an input query $x$ and a 
\ptdyI{response} $y$ to a scalar score.
In order to do that, we associate each feature $c$ with a specific prompt $t_c$ and compute a score $r_c=R_c(y|x,t_c)$, where $R_c$ can be a general LLM like GPT-3.5, prompted \reb{with a combination of $t_c$, $x$, and $y$}.
\ptmdII{These} features are designed to decompose the broad concept of preferability into  a series of more straightforward and interpretable components.\footnote{\reb{See \citet{sharma2023understanding} and \citet{hosking2023human} for further evidence that human preference judgements can be accurately predicted from a linear combinations of such features.}}  \reb{\ptdyII{In general, the features should be ``diverse" enough so that they can cover the broad concept of preference, 
yet without too much ``overlap" between them to decrease  efficiency and interpretability.}} It is noteworthy that a feature can 
represent
not only positive categories that are aligned with preferability (e.g. informativeness), but also categories that are assumed to be negatively correlated with it
(e.g. \ptdyI{biased\ptmdII{ness}}). 
This procedure allows us to control which properties of the preference data are used to train the PM and 
to build \ptmdII{it} based on \ptmdII{components} that we believe to 
{determine}
the human choices.

\subsection{Combining multiple features}\label{sec:combining_features}

The features assessed by the prompted LM
serve as distinct modules, each of which evaluates a different aspect. 
\ptdyI{To combine the features into an interpretable single model,
we employ logistic regression to classify the preferred response in \ptmdII{a} pairwise comparison dataset.}\footnote{Expanding pairwise comparisons to rank data is possible, following \ptmdI{the} general approach of one-vs-one \citep{ouyang2022training}.}

\reb{Based on the dataset $\mathit{\mathcal{D}}=\{x^{i},y_{1}^{i},y_{2}^{i}\}_{i=1}^{N}$, we obtain a feature matrix $\{x^{i},\boldsymbol{r}(y_{1}^{i}|x^{i}),\boldsymbol{r}(y_{2}^{i}|x^{i})\}_{i=1}^{N}$. 
Here $\boldsymbol{r}(y|x)=(R_1(y|x,t_1),\ldots,R_C(y|x,t_C))$ is a  feature vector \ptdyIII{with} %
decomposed feature scores.
We standardize each feature score to have average $0$ and variance $1$ within the train data.
We \ptdyIII{then} compute the pairwise difference of the feature vectors for each pair of responses, $\boldsymbol{r}(y_{1}|x)-\boldsymbol{r}(y_{2}|x)$, and train a logistic regression classifier with this difference 
to predict $1$ if $y_1$ is preferred, and $0$ if $y_2$ is preferred. In other words, the distribution $p$ is formalized as:
\begin{equation}\label{eq:CPM}
p(y_{1}>y_{2}|x)\doteq \sigma(\left\langle \boldsymbol{\lambda},\boldsymbol{r}(y_{1}|x)-\boldsymbol{r}(y_{2}|x)\right\rangle)=(1+\exp(\left\langle \boldsymbol{\lambda},\boldsymbol{r}(y_{2}|x)-\boldsymbol{r}(y_{1}|x)\right\rangle))^{-1}
\end{equation}\fmdIII{For comparability with Eq\ref{eq:standard-PM}}
\ptdyI{where $\boldsymbol{\lambda}=(\lambda_1,\ldots,\lambda_C)$ is \ptmdI{the} vector of fitted coefficients}. 
The coefficient $\lambda_c$ indicates the importance of the feature $c$ for predicting human preference judgements. To obtain the preference score of a single sample we simply compute 
$\left\langle \boldsymbol{\lambda},\boldsymbol{r}(y|x) - \boldsymbol{0} \right\rangle = 
\left\langle \boldsymbol{\lambda},\boldsymbol{r}(y|x)\right\rangle$, where $\boldsymbol{0}$ is the standardized average of the feature vector $\boldsymbol{r}(y|x)$ over the training data as explained above.
} %

\section{Experiments}

In this section, we empirically evaluate 
CPM on \ptmdII{several} aspects, including model robustness (Sec.~\ref{sec:model_robustness}), generalization (Sec.~\ref{sec:referece_comparison}), robustness to overoptimization (Sec.~\ref{sec:overoptimization}), and effectiveness for preference alignment (Sec.~\ref{sec:quality_eval}).
We also provide an illustrative example of CPM \ptmdII{interpretability} in Sec.~\ref{sec:interpretability}.

\subsection{Experimental setup}

\paragraph{Datasets}
We conduct experiments on two datasets, the HH-RLHF dataset \citep{bai2022training} and \ptmdII{the} SHP dataset \citep{shpdata}. Both consist of pairs of responses based on helpfulness. For each dataset, 
\ptdyII{in order to establish a consistent setting and control for the data size factor, we sample 20K single-turn data points.}
\vspace{-2px}
\paragraph{Features}
We use 13 features\ptmdI{:} 
\texttt{helpfulness, specificity, intent, factuality, easy-to-understand, relevance, readability, enough-detail, biased, fail-to-consider-individual-preferences, repetitive, fail-to-consider-context} and \texttt{too-long},
with pre-specified prompt  templates (see App.~\ref{app:prompts} for the description of features and prompts). We use the same set of features for both datasets; prompt templates only differ in a preamble that describes $x$ as either a conversation with \ptmdI{an} AI assistant (HH-RLHF) or a StackExchange question (SHP).
We also use the length of $y$\ptmdII{,} which we find to be helpful 
on the SHP dataset. 
\vspace{-3px}
\paragraph{Methods}

To find out the ability of an LM as a feature extractor,
we explore two LMs, GPT-3.5
(\texttt{gpt-3.5-turbo-0301}) and Flan-T5-XL (3B parameters) \citep{flant5}, using the same features and prompt templates. \ptdyI{We refer to \ptmdII{the CPM models based on these extractors}  as CPM-GPT-3.5 and CPM-Flan-T5, respectively.}
To select only the most important features, we add a 
regularization term in logistic regression and use hyperparameters selected with 5-fold cross-validation on the training dataset.

We \ptmdII{then} compare \ptmdII{the} conventional PM to \ptmdII{these} CPMs (trained respectively as described in Sec.~\ref{sec:preliminaries} and Sec.~\ref{sec:combining_features}). 
\ptdyI{For a fair comparison, we train the standard PM \ptmdII{based on} the same Flan-T5-XL model that we use for 
\ptdyII{the}
CPM\ptmdII{s}, but with an added linear head that outputs a scalar preference score.}
We compare the performances of CPM-GPT-3.5 \ptmdI{and} CPM-Flan-T5 with 
\ptmdII{this} standard PM.
Implementation details %
are provided in App.~\ref{app:hyperparams}.
\vspace{-3px}
\paragraph{Best-of-$n$ sampling (BoN)}

To assess the robustness of PMs to overfitting, we use 
Best-of-$n$ 
(BoN) sampling \citep{gao2023scaling},
\ptdyI{a simple yet effective method that has been shown to be competitive with more advanced techniques such as reinforcement learning \citep{openaibon}.} 
\ptdyI{BoN abstracts away from RLHF design choices such as the details of policy optimization and provides a stable proxy for RLHF performance \citep{nakano2021webgpt,gao2023scaling}.}

We generate $n$ responses using an initial LM 
$a(x)$ and evaluate the performance of the PMs on these responses.\fbtkI{There are two things IMHO: computing PMs scores to rank responses and computing PMs scores to evaluate models \ptdyI{DY: (25/09/2023) Could you elaborate more on this please?} It soundes a bit confusing and wanted to suggest a rewrite (mentioning that PM is also used in generating BoN samples) but actually maybe it's okay as is?} We consider the BoN distribution $x\sim\text{BoN}(a,\text{PM},n)$, where $n$ candidates are sampled from $a$ and $x$ is the candidate maximizing the PM score.
\ptdyI{Following \cite{gao2023scaling},} we compare the robustness of two related PMs, $\text{PM}_{A}(x)$ and $\text{PM}_{B}(x)$, 
by \ptdyI{measuring the gap} between their \ptmdI{average scores relative to samples $x$ from $\text{BoN}(a,\text{PM}_A,n)$, where typically (by construction) 
we have $\text{PM}_A(x) > \text{PM}_B(x)$, with the gap increasing with $n$.}%
\footnote{The PM used for the BoN distribution is determined by the experimental design (e.g. proxy PM in the overoptimization experiment).}
%
%

%

We generate up to 25,600 BoN responses, with 256 responses for each of 100 prompts in \ptmdI{a} held-out test set.\footnote{Due to computational constraints, 
\ptdyII{we only evaluate CPM-GPT-3.5 on 
BoN($n\le 16$).
}
}
We use Flan-T5-Large \citep[780M parameters;][]{flant5} as the initial LM to generate the responses. 
To ensure that the performance of different PMs can be compared 
on the same scale
across different reward models, we normalize each PM score to have average 0 and variance 1 within the training data.
%
%

%

%

%
%

\subsection{Model robustness}\label{sec:model_robustness}

\begin{figure}[ht]
\centering
\begin{tabular}{cc}
(a) HH-RLHF dataset\\
\centerline{\includegraphics[width=\columnwidth]{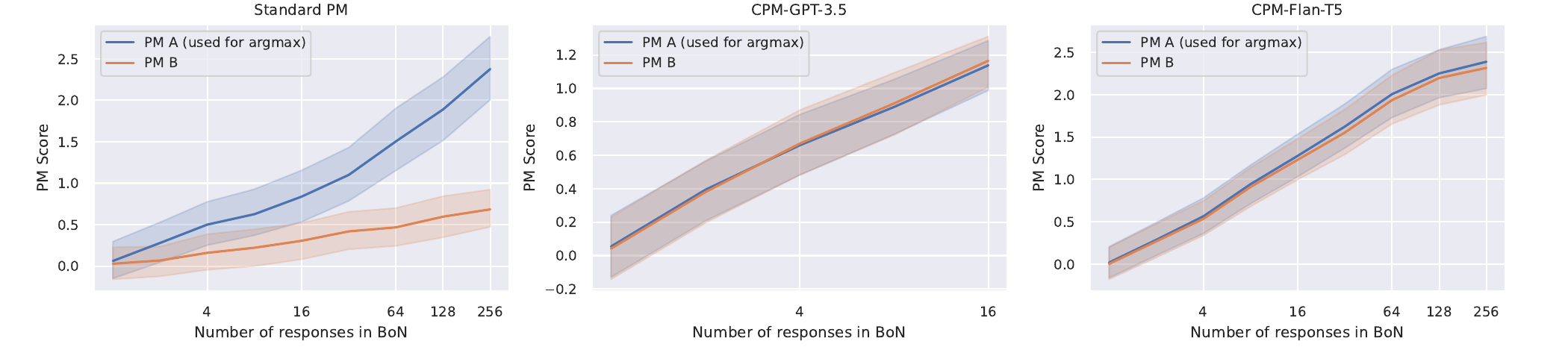}}\\
(b) SHP dataset\\
\centerline{\includegraphics[width=\columnwidth]{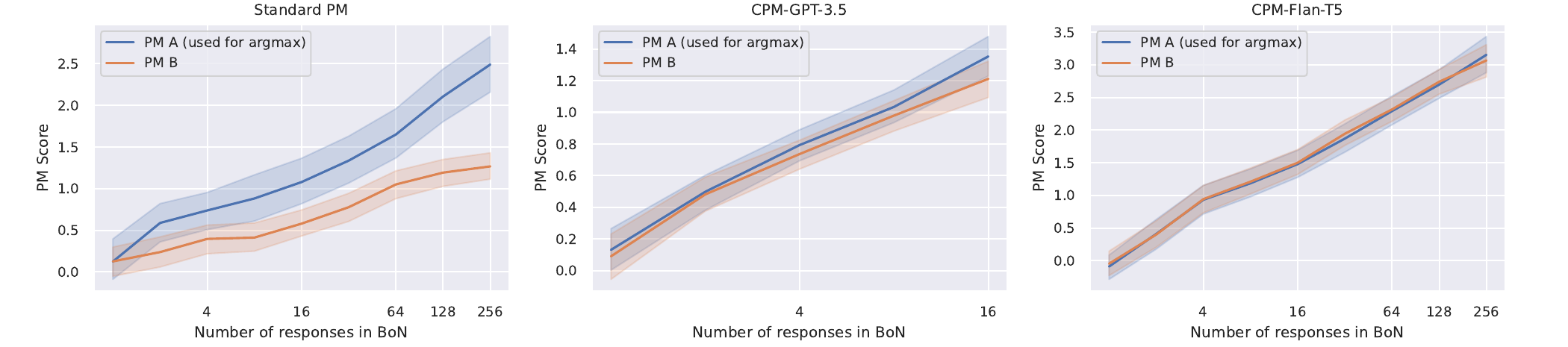}}
\end{tabular}
\caption{
BoN comparison over two models fitted independently in same condition \reb{(left: Standard PM, middle: CPM-GPT-3.5, right: CPM-Flan-T5)}\ptmdI{.} 
\ptdyIII{PM A (blue line) is used for BoN selection.}
}\label{fig:model-variance}
\vspace{-5px}
\end{figure}

Model robustness refers to the sensitivity of a predictive model to the selection of its training data \citep{hastie2009elements}.
Specifically, it quantifies how much the model's predictions would change if we were to train it on different subsets of the preference dataset.
A model with low robustness
will show poor generalization on unseen data.

To assess model robustness, we independently train two  PMs for each PM method\ptdyI{, $\text{PM}_{A}$ and $\text{PM}_{B}$,} on disjoint subsets of the training data\ptdyI{, each of size 10K.}
We then conduct a BoN experiment and check whether the scores of these two PMs diverge with increasing $n$.
\ptmdII{As explained above}, we pick the response with highest $\text{PM}_{A}$ score among $n$ samples and measure the gap between the scores of $\text{PM}_{A}$ and $\text{PM}_{B}$ on that sample.\footnote{We tested reversing the order for building BoN distribution, and the results remained unchanged. See Fig.~\ref{fig:model-variance-app} in \ptmdI{the} Appendix.}

Fig.~\ref{fig:model-variance} shows that CPM
is significantly more consistent between  $\text{PM}_{A}$ and $\text{PM}_{B}$
than the standard PM method in terms of 
\ptdyI{the score differences, }
even for BoN with size $256$.
The smooth 
scaling trend as a function of $n$ suggests that our findings will 
\ptmdI{generalize} to larger $n$. This suggests that the small number of trainable coefficients (in this experiment 14 coefficients) makes the model robust to \ptdyI{noise in data sampling.} 
\ptmdI{Still,}
the features extracted by LM are informative enough to 
build an effective preference model for alignment tuning, as we \ptmdII{illustrate below.}

\subsection{Comparison with reference PM\ptmdI{s}}\label{sec:referece_comparison}%

\begin{figure}[ht]
\begin{centering}
\begin{tabular}{cc}
{\includegraphics[width=0.48\columnwidth]{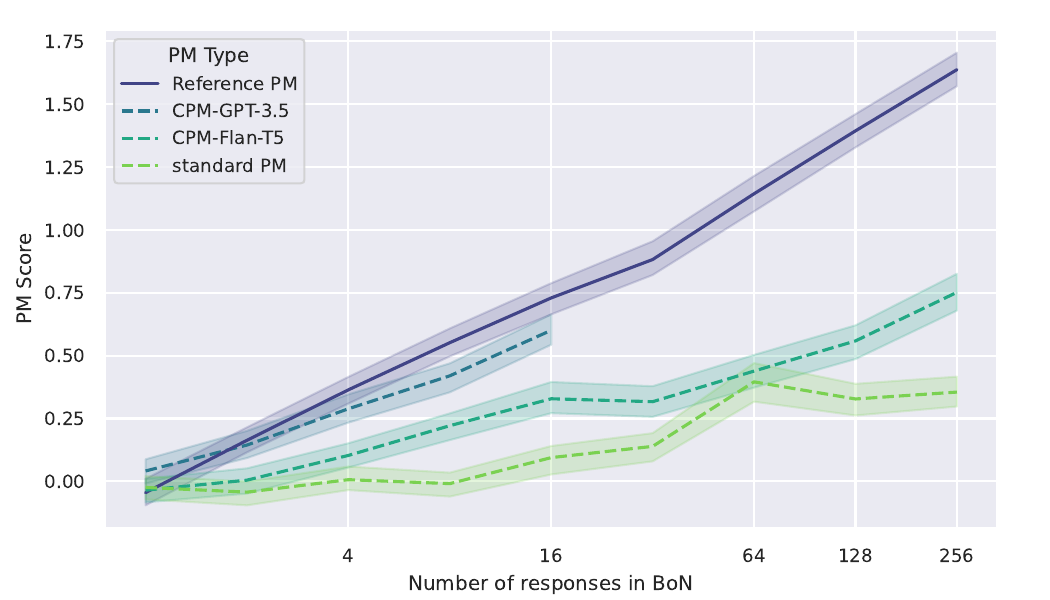}}  & {\includegraphics[width=0.48\columnwidth]{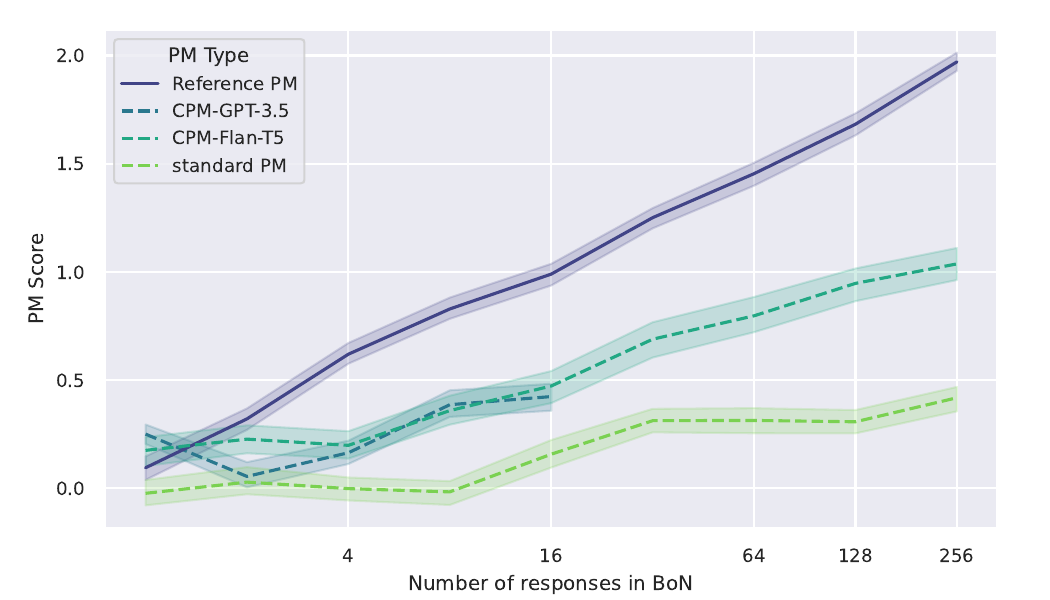}}  \tabularnewline
\end{tabular}
\par\end{centering}
\caption{
\ptdyI{Comparison \ptmdII{between} PM scores \ptmdII{relative to the distributions} $\text{BoN}(a,\text{PM}_\text{ref1},n)$ (HH-RLHF dataset, left) and   $\text{BoN}(a,\text{PM}_\text{ref2},n)$ (SHP-dataset, right). }
}\label{fig:external_comparison}
\end{figure}

\ptdyII{To assess the generalizability
of our CPM\ptmdII{s}, we compare \ptmdII{them} to two well-established reference PMs, $\text{PM}_{\text{ref}1}$ and $\text{PM}_{\text{ref}2}$, both instances of DeBERTa \citep{he2020deberta}, with $\text{PM}_{\text{ref}1}$ finetuned on a large dataset including HH-RLHF\footnote{\href{https://huggingface.co/OpenAssistant/reward-model-deberta-v3-large-v2}{https://huggingface.co/OpenAssistant/reward-model-deberta-v3-large-v2}} and $\text{PM}_{\text{ref}2}$ finetuned on a large dataset including SHP \citep{sileo2023tasksource}.}
These 
PMs, trained on larger and more diverse datasets,
are shown to generalize better \ptdyI{than PMs trained on a 10K dataset}
(see App.~\ref{app:deberta_evaluation}).
We select BoN responses with the reference PM and then examine how their scores diverge 
relative to
\ptmdII{the different PMs} trained on a 10K dataset as in Sec. \ref{sec:model_robustness}.
\ptdyII{We hypothesize that models that diverge less from such independently
trained reference PMs will generalize better to unseen data.}
Fig.~\ref{fig:external_comparison}
\ptdyI{shows that all models  
\ptmdII{scale monotonically}
with the reference PM,}
with \ptmdII{the} CPM\ptmdII{s} \ptmdII{staying} %
closer to it.
This suggests that \ptmdII{the extracted} features 
are informative enough to allow for learning a 
\ptmdII{more generalizable}
model of preference judgements.

\subsection{Robustness to Overoptimization}\label{sec:overoptimization}

\begin{figure}[ht]
\begin{centering}
\begin{tabular}{cc}
{\includegraphics[width=0.48\columnwidth]{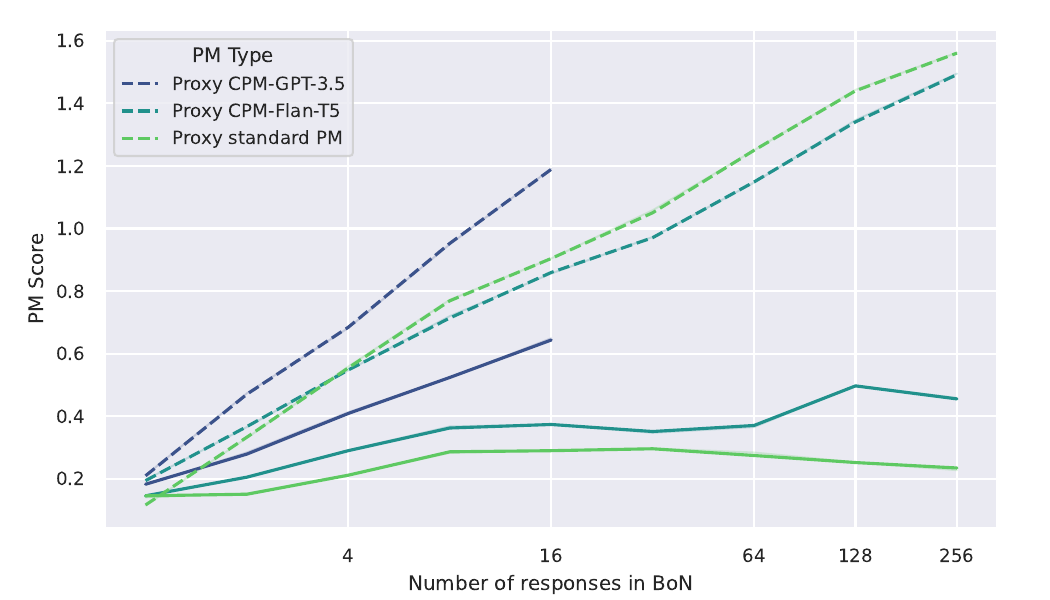}}  & {\includegraphics[width=0.48\columnwidth]{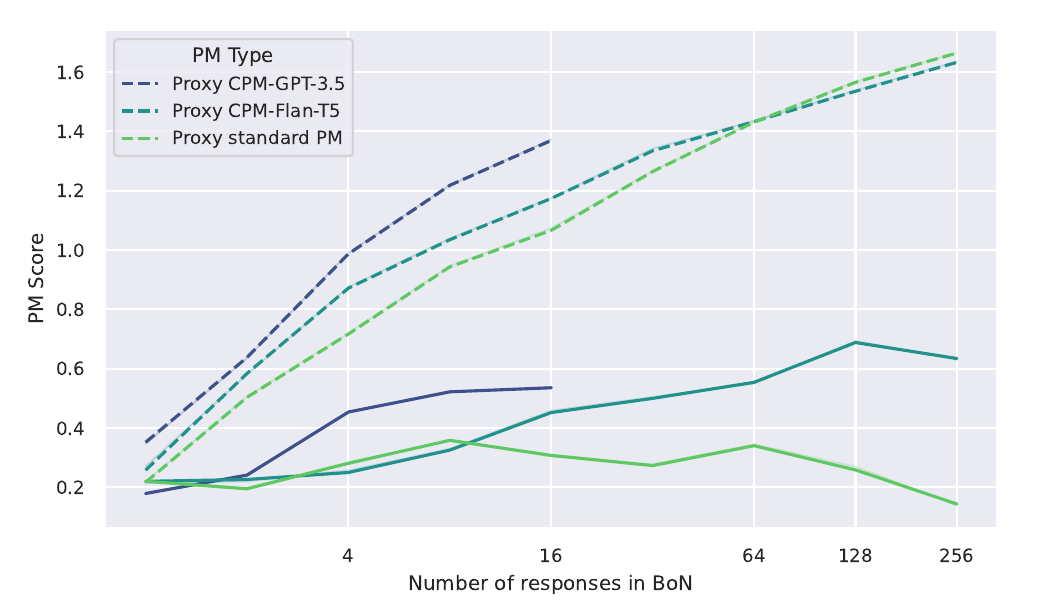}}  \tabularnewline
\end{tabular}
\par\end{centering}
\caption{Overoptimization experiment in BoN distribution $\text{BoN}(a,\text{PM}_\text{Proxy},n)$. 
\ptdyII{Dashed line means proxy PM used for BoN selection, corresponding solid line means gold PM.}
(left: HH-RLHF dataset, right: SHP dataset)
\label{fig:overeoptimization}}
\end{figure}

Overoptimization is a type of misalignment that occurs when the 
{preference model}
is overly optimized by exploiting flaws in the proxy objective \citep{amodei2016concrete,skalse2022}.
This can lead to the PM diverging from the true objective, which we want to 
\ptmdII{optimize}
in alignment tuning. 

To investigate overoptimization, we follow \cite{gao2023scaling} and construct a \ptmdI{synthetic} dataset where the output of \ptmdI{a} specific \ptmdI{``gold''} PM is \ptmdI{assumed} to be the ground truth \ptmdI{for preferences}. 
\ptmdI{As gold PMs}, we use reference PMs \ptdyI{$\text{PM}_{\text{ref}1}$ and $\text{PM}_{\text{ref}2}$} (described in Sec. \ref{sec:referece_comparison}). 
\ptdyII{We then use the gold models to generate synthetic labels to train proxy PMs using each of the studied techniques.}
\ptmdI{Depending on the PM training method, overoptimizing the PM can cause it to diverge from the gold PM, which allows us to compare the robustness of different PM techniques.}

Fig. \ref{fig:overeoptimization} shows that the gap between the gold PM and \ptmdI{the} proxy PM scores increases for each PM as the candidate size $n$ increases. 
The distribution of the standard PM does not follow the gold PM distribution and has \ptmdI{a} larger divergence as \ptmdI{the} candidate size $n$ increases. 
This 
\ptmdI{illustrates}
that 
{fitting}
{a} standard PM 
can lead to overoptimization, which is consistent with existing literature \citep{gao2023scaling}. %
On the other hand, 
\ptdyI{the gap between \ptmdI{the} gold and proxy PM scores is smaller for 
CPMs, \ptmdI{with the} gold PM score \ptmdI{beginning} to diverge later than for standard PMs. This suggests that CPMs are more robust to overoptimization.}
The \ptdyI{rank correlation of the PM scores with increasing $n$ in Fig. \ref{fig:overeoptimization},}
which measures this quantitatively, is provided in Table ~\ref{tab:overoptim_corr} \ptmdI{in the Appendix}.

\subsection{Quality evaluation}\label{sec:quality_eval}

The ultimate goal of PMs is to help align LMs with human preferences. While in the previous section we compared PMs with a certain gold PM, in this section we will investigate whether LMs aligned using CPMs are preferred by humans over LMs aligned using standard PMs. 
Following previous literature \citep{vicuna2023,mukherjee2023orca,liu2023gpteval,he2023annollm}, we simulate human evaluation using a prompted \ptmdI{LLM.}

For each PM, we 
\ptdyII{draw}
\ptmdII{a response} from $\text{BoN}(a,\text{PM},16)$ by generating \ptmdI{samples} 
from 
\ptdyII{$a$ \ptmdII{(namely Flan-T5)} }
and select\ptmdI{ing the} best \ptmdI{response} based on \ptmdI{the} PM score.
We then compare \ptmdI{this} response to vanilla Flan-T5, \ptmdI{namely a} response randomly selected from the same set of candidates.
We \ptmdII{finally} use 
\ptdyII{the LLM}
to 
\ptmdII{choose which response}
is 
preferable.
We refer to this metric as the ``win rate''. 
A good PM is expected to have high win rate against vanilla Flan-T5. 

Importantly, 
we use Claude \citep[\texttt{claude-2};][]{claude}, an LLM that
was \emph{not} used in feature extraction. Hence, we avoid \emph{potential }subtle preference leaks from features extracted usig GPT-3.5\ptmdI{.} We use the prompt from \citep{vicuna2023,mukherjee2023orca} to rate the quality of \ptmdI{the} response selected by each PM method\footnote{To prevent the known bias towards the first 
response \citep{vicuna2023,openai2023gpt4},  we average the scores with different orderings when making a comparison.} (see Tab.~\ref{tab:claude_prompt} for the prompt used in evaluation).
We perform one BoN trial with $n=16$ for CPM-GPT-3.5 and 10 independent \ptmdI{such} trials for other PMs and report the average win rate.

\begin{wraptable}{t}{0.5\textwidth}
\begin{centering}
\vspace{-10px}
\begin{tabular}{c|cc}
\toprule
Win Rate & \multicolumn{1}{c}{HH-RLHF} & \multicolumn{1}{c}{SHP}\tabularnewline
\midrule
CPM-GPT-3.5 & \textbf{0.810}\quad\quad(.) & \textbf{0.672}\quad\quad(.)\tabularnewline
CPM-Flan-T5 & 0.742 (0.034) & 0.580 (0.045)\tabularnewline
Standard PM & 0.588 (0.030) & 0.564 (0.037)\tabularnewline
\bottomrule
\end{tabular}
\par\end{centering}
\caption{
Win rate over initial generation after BoN sampling based on each PM. 
Except CPM-GPT-3.5, we independently conduct $10$ rounds of BoN($n=16$) samplings and report the average win rate along with standard error.\label{tab:claude_eval}
\vspace{-10px}
}
\end{wraptable}

Tab.~\ref{tab:claude_eval} shows evaluation results. Considering that both standard PM and CPM-Flan-T5 use the same architecture \ptdyI{ and data}, 
the higher win rate of CPM-Flan-T5 compared to standard PM suggests the 
\ptmdI{advantage}
of decomposing preference into multiple features and using an LM as feature extractor, 
rather than directly using the PM based on fine-tuning the LM as in Eq. (\ref{eq:standard-PM}).
CPM-GPT-3.5 shows 
\ptmdI{an even higher win rate, again indicating} %
that using a more powerful 
\ptmdI{LM}
as feature extractor can further improve the performance of CPM.

\subsection{Model interpretability}\label{sec:interpretability}

CPM\ptmdI{s}, as linear models, 
\ptmdI{have a high degree of interpretability}
\cite{hastie2009elements}.
In this section, we provide a few illustrative examples \ptmdII{focussing on the dataset HH-RLHF}.

\paragraph{Coefficients}
The interpretability of our model is enhanced by the fact that the feature coefficients provide a direct indication of the factors that most influence the CPM's decisions. This information can \ptmdII{help} understand the CPM's internal workings. Tab.~\ref{tab:fitted_coeff}
shows \ptmdI{the} top 3 largest coefficients
(see Tab.~\ref{tab:full_fitted_coeff} for full coefficients). Although the coefficient\ptmdI{s} var\ptmdI{y} as they are extracted with different LMs, 
\ptmdI{their orders are generally consistent, }
except for a few features. This 
\ptmdI{observation}
provides some clues into how the CPM makes its decisions. In the current 
\ptdyI{example,}
the CPM focuses on general helpfulness and also prefers response\ptmdI{s} that are detailed enough but also factually correct.

\begin{table}[htb]
\vspace{-5px}
\begin{centering}
\begin{tabular}{cccc}
\toprule
\multicolumn{2}{c}{CPM-GPT-3.5} & \multicolumn{2}{c}{CPM-Flan-T5}\tabularnewline
\cmidrule(lr){1-2}\cmidrule(lr){3-4}
Feature & Coefficient & Feature & Coefficient\tabularnewline
\midrule
helpfulness & 0.246 & fail-to-consider-context & 0.420\tabularnewline
enough-detail & 0.235 & enough-detail & 0.244\tabularnewline
factuality & 0.187 & factuality & 0.227\tabularnewline
\bottomrule
\end{tabular}
\par\end{centering}
\caption{%
\ptmdI{Three largest CPM coefficients}
on HH-RLHF dataset. 
\label{tab:fitted_coeff}}
\vspace{-5px}
\end{table}

\paragraph{LM-extracted features}
The features extracted by the LM enable intuitive explanation of generated responses. This 
allows
supervising complex behavior in \ptmdI{a} human-interpretable way.  Tab.~\ref{tab:feature_values}
shows examples of these features,
which can be used to identify which aspects of the response contribute most to the predicted preference judgement.  By decompos\ptmdI{ing} a hard preference (``This text is not preferable.”) into a series of easier features (``This text is generally unhelpful, as it is easy to read but has little detailed information"), it allows \ptmdI{easier inspection}
for a human overseer.

\begin{table}[htb]
\vspace{-5px}
\centering{}%
\begin{tabular}{lccccc}
\toprule
    & \multicolumn{5}{c}{\footnotesize{}Features}\tabularnewline
\multicolumn{1}{c}{\footnotesize{}Dialogue} & {\tiny{}\rotatebox[origin=c]{90}{helpfulness}} & {\tiny{}\rotatebox[origin=c]{90}{readability}} & {\tiny{}\rotatebox[origin=c]{90}{enough-detail}} & {\tiny{}\rotatebox[origin=c]{90}{factuality}} & {\tiny{}\rotatebox[origin=c]{90}{\makecell{fail-to-consider-\\context}}}\tabularnewline
\midrule
{\tiny{}Q: What things can I create with java programming?} \tabularnewline
    
{\makecell[l]{\tiny{}R1: Java is a programming language that is designed to create applications that run on a computer.}} & {\tiny{}1.209} & {\tiny{}0.559} & {\tiny{}0.913} & {\tiny{}1.690} & {\tiny{}2.117}\tabularnewline
    
{\tiny{}R2: Robots} & {\tiny{}0.509} & {\tiny{}0.559} & {\tiny{}-0.075} & {\tiny{}-0.569} & {\tiny{}-0.239}\tabularnewline

\midrule
{\tiny{}Q: Why is Lucifer referred to as the ``morning
star"?} \tabularnewline
    
\makecell[l]{\tiny{}R1: Lucifer is the Morning Star because it is the brightest
star in the sky.} & {\tiny{}1.209} & {\tiny{}0.880} & {\tiny{}1.901} & {\tiny{}0.843} & {\tiny{}0.232}\tabularnewline
    
{\tiny{}R2: angel} & {\tiny{}-0.890} & {\tiny{}0.239} & {\tiny{}-0.569} & {\tiny{}-0.851} & {\tiny{}-0.239}\tabularnewline

\midrule
{\tiny{}Q: I'd like to evaluate some art perhaps. Can you tell me some history about Mona Lisa, the painting?} \tabularnewline
    
{\makecell[l]{\tiny{}R1: The Mona Lisa is a famous painting in the Louvre, Paris, France. It is attributed to Leonardo da Vinci.}} & {\tiny{}1.909} & {\tiny{}0.880} & {\tiny{}1.901} & {\tiny{}1.690} & {\tiny{}2.588}\tabularnewline
    
{\makecell[l]{\tiny{}R2: It was painted by Gian Lorenzo da Sica in the late 18th century. The name of the painting is inspired by \\\tiny{}the story of the famous Italian painter Leonardo da Vinci. The painting has been described as having a\\\tiny{} ``poetic"  quality. }} & {\tiny{}0.859} & {\tiny{}0.239} & {\tiny{}1.901} & {\tiny{}0.278} & {\tiny{}-0.239}\tabularnewline

\bottomrule
\end{tabular}\caption{Examples of feature values of CPM-GPT-3.5. 
Each feature value can provide information about which aspects of the response are good or bad. \label{tab:feature_values}}
\vspace{-1px}
\end{table}

\section{Ablation Study}
\label{sec:ablation}

In \ptmdI{this} ablation study, we evaluate the impact of two factors, data size and size of the LM for feature extraction on \ptmdII{the} HH-RLHF dataset. 

\begin{figure}[ht]
\vspace{-5px}
\begin{centering}
\begin{tabular}{cc}
{\includegraphics[width=\columnwidth]{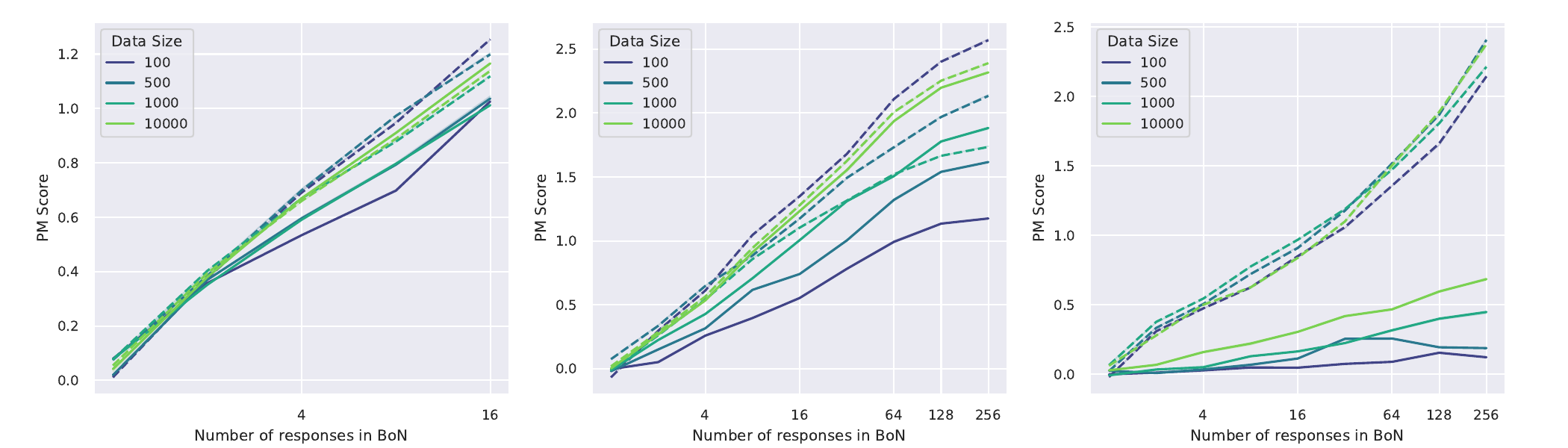}} 
\tabularnewline
\end{tabular}
\par\end{centering}
\caption{
BoN comparison of two models fitted independently with scaling data size in HH-RLHF dataset (left: CPM-GPT-3.5, middle: CPM-Flan-T5, right: standard PM\ptmdI{)}.}\label{fig:hh_data_scaling}
\end{figure}

\vspace{-3px}
\paragraph{Data scaling trend}
To investigate the effect of data size on model robustness, we hold the model size constant (3B parameters) and vary the data size used to train the PMs.
We independently train 
\ptmdI{each PM method on two}
disjoint subsets of the training data, as described in Sec.~\ref{sec:model_robustness}. We gradually increase the data size from 100 to 10,000.
Fig.~\ref{fig:hh_data_scaling} shows the results of the model robustness experiment. CPMs rapidly become consistent as the data size increases and achieve stable consistency between two PMs with a data size of over 500. In contrast, standard PMs show poor consistency between models, especially when the data size is small. This suggests that CPMs are more robust than standard PMs and can produce reliable results even with a small amount of data.

\begin{figure}[ht]
\begin{centering}
\begin{tabular}{ccc}
{\includegraphics[width=0.5\columnwidth]{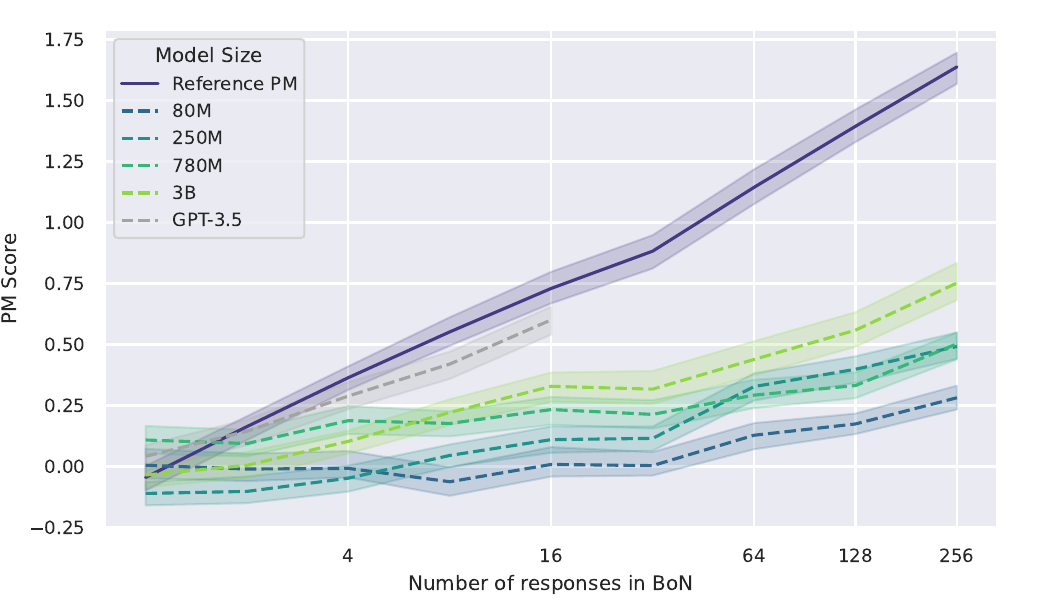}} & {\includegraphics[width=0.5\columnwidth]{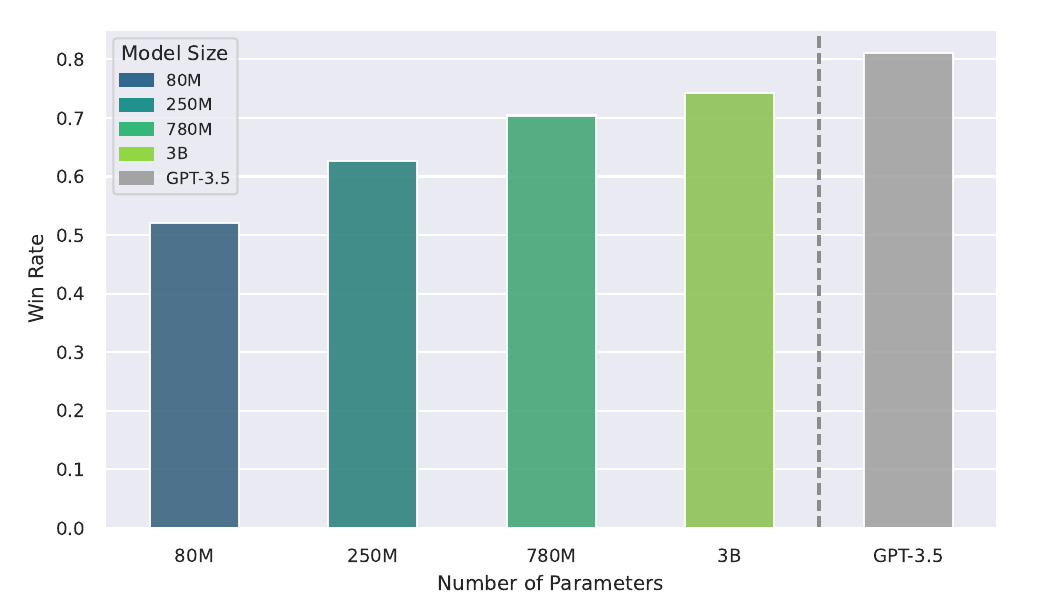}} \\ 
 \tabularnewline
\end{tabular}
\par\end{centering}
\vspace{-20px}
\caption{
Model size scaling experiment using Flan-T5. 
(left: comparison with the reference PM, right: win rate over initial generation after BoN sampling based on each PM)
\vspace{-8px}
}\label{fig:hh_model_scaling}
\end{figure}

\vspace{-3px}
\paragraph{Model scaling trend}
To investigate the effect of \ptmdI{the size of the LM}
used for feature extraction, 
we gradually increase \ptmdI{this} size from Flan-T5 ``small'' (80M parameters) to ``XL'' (3B parameters) and track two important metrics: model generalizability (described in Sec.~\ref{sec:referece_comparison}) and win rate (described in Sec.~\ref{sec:quality_eval}). The training data size is fixed to 10K.
As shown in Fig.~\ref{fig:hh_model_scaling},
both model generalizability and win rate steadily improve with \ptmdI{increasing} LM size. This confirms that LM capability propagates to feature extraction, and that CPM can take advantage of it. 
This further means that CPM\ptmdII{s} can become even more useful as \ptmdII{extractor} LMs become more capable. The smooth and gradual increase of the win rate as a function of LM size suggests that our findings generalize to the case of using even larger LMs for feature extraction.

\section{Related work}

\paragraph{Robustness of preference models}
PM overoptimization is an instance of reward hacking, a situation when a policy exploits flaws in its reward function \citep{amodei2016concrete,skalse2022}. These flaws can come from errors of human evaluators  \citep{pandey2022modeling}, the inherent difficulty of learning preferences of irrational agents \citep{mindermann2018occam,shah2019feasibility} or \ptmdI{the} fragility of learned reward functions \ptmdI{to} adversarial attacks \citep{mckinney2023fragility}. \cite{gao2023scaling} studied the scaling properties of PM overoptimization and \cite{casper2023open} discuss it in a broader context of open problems with RLHF. More generally, PMs can learn to be sensitive to spurious features associated with human feedback. This leads to failure modes such as sycophancy \cite[a tendency to answer a question with a user's preferred answer, even if that answer is not correct;][]{cotra2021why,perez2022discovering} or social bias \citep[due narrow demographics of feedback providers;][]{santurkar2023whose,hartmann2023political}. Despite its growing importance, the problem of learning robust PMs for aligning LMs is largely neglected. The present paper attempts to fill this gap.

\vspace{-1px}
\paragraph{Decomposing tasks for LMs.}
There are numerous examples of task decomposition increasing the accuracy or robustness of language models. 
Breaking down problems into steps \citep[chain-of-thought;]{wei_chain--thought_2022} or into a sequence of subproblems depending on answers to previous subproblems \citep{zhou2023leasttomost} are enormously beneficial for tasks involving reasoning. 
Others explored a stronger separation: solving subproblems independently in different LM context windows. For instance, 
\cite{creswell2022selectioninference} alternate between selection and inference to generate a series of interpretable, casual reasoning steps. 
\cite{radhakrishnan2023question} found that solving subproblems in separate context windows improves faithfulness of reasoning. 
\cite{reppert2023iterated} build compositional LM programs by applying decomposition iteratively, with a human in the loop, to facilitate science question answering. The present paper find\ptmdI{s} similar robustness benefits of decomposition for preference modeling.
\vspace{-1px}
\paragraph{Scalable oversight}
Scalable oversight is the problem of evaluating the behaviour of agents more capable than the evaluators \citep{bowman_measuring_2022}. On the one hand, LMs may
soon grow capable of completing tasks for which humans will not be able to provide feedback. On the other, LMs might also be capable of reasoning about flaws in their evaluation procedures \citep{berglund2023taken} and exploiting them unbeknownst to overseers. Current proposals for solving scalable oversight focus on recursively relying on other LMs to assist human evaluators \citep{irving2018ai,leike2018scalable,christiano2018supervising}. RL from AI feedback \citep{bai2022constitutional} attempts to implement this idea by using carefully prompted LMs to generate training data for PMs. In contrast, we propose to rely on LMs 
during a single inference step of a PM.

\section{Conclusion}

We introduce Compositional Preference Models (CPMs), a simple and effective paradigm for training robust and interpretable preference models. CPMs decompose global preference scores into interpretable features and rely on language models (LMs) to extract those features. Despite their simplicity, CPMs are robust to different \ptmdII{sub}samplings of the dataset and to overoptimization, and they outperform conventional preference models at obtaining preferred best-of-$n$ samples. We believe that CPMs pave the way for combining human insights into preference judgements with the LM capabilities to extract them. Given the recent advances in LM \ptmdI{abilities}, CPMs have the potential to be\ptmdII{ing} used for alignment and scalable oversight of \ptmdII{models} with superhuman capabilities. 
\reb{One limitation of our work is that instead of a genuine human evaluation of the preferences, 
we use a proxy LLM (Claude 2) for \ptdyIII{the} evaluation. 
One research direction here could be to introduce a task-oriented generation scenario \ptdyIII{(e.g. task accomplishment)} where helpfulness 
could be evaluated easily and to understand how 
to inform the preference model with this scenario.}
%
%
\ptmdII{Finally, another possible objective for future research would be to explore how to \emph{elicit} decomposed features that can capture various kinds of complex preference judgements. A promising direction here would be to leverage LMs to not only score, but actually \emph{discover} the component features that determine these judgements.

}%

%
%
%
%
%
%
%

\clearpage


\appendix
\clearpage

\section{Implementation Details}\label{app:hyperparams}

\subsection{Compositional preference model}\label{app:cpm_hyperparams}

We used GPT-3.5 (\texttt{gpt-3.5-turbo-0301}) and Flan-T5-XL (3B parameters) \citep{flant5} as a feature extractor, using the same features and prompt templates in Tab.~\ref{tab:hh-prompt} and Tab.~\ref{tab:shp-prompt}. We excluded randomness from the generation process and selected the token with the highest likelihood.

For logistic regression classifier we used Scikit-learn \citep{sklearn_api}. We set the choice of $L_1$ and $L_2$ regularization, weight of regularization, and solver of the logistic regression classifier as a hyperparameters and selected best hyperparameters based on 5-fold cross-validation in training dataset.

In the inference time, we made feature scores of the generated response using same LLM and templates used in training phrase. The feature scores are aggregated  with the trained logistic regression classifier as described in Sec.~\ref{sec:combining_features}.

\subsection{Standard preference model}\label{app:pm_hyperparams}

All standard PMs were implemented using PyTorch \citep{paszke2019pytorch} and HuggingFace Transformers \citep{wolf2020transformers} We adopt the AdamW optimizer \citep{loshchilov2017decoupled} with $\beta = (0.9, 0.98)$ and set the weight decay to $0.01$. 
We conducted separate hyperparameter sweeps over learning rate and batch size for each dataset, using early-stopping based on the evaluation set with 3 steps of patience.
We used a batch size of 32 and a learning
rate of 1e-5 for HH-RLHF dataset and 5e-5 for SHP dataset. We used cosine learning rate schedule with 100 linear warmup steps. 
We used Flan-T5-XL \citep[3B parameters]{flant5} for standard PMs, which is available  on the Huggingface Model Hub under the model name of \texttt{google/flan-t5-xl}.  
Training was performed on Nvidia A100 GPU, with the longest run taking approximately 12 hours. 

\section{Claude evaluation of  the reference PM}\label{app:deberta_evaluation}

To evaluate the performance of reference PM in Sec.\ref{sec:referece_comparison} in preference alignment, we follow the same quality evaluation framework as in Sec.~\ref{sec:quality_eval}. Specifically, we select the best sample among 16 responses generated by Flan-T5, based on the reference PM score. 
We then compare {this} response to vanilla Flan-T5, a response randomly selected from the same set of candidates,  as described in Sec.~\ref{sec:quality_eval}.

Again, we use Claude 
to rate the quality of the response selected by reference PMs (see Tab.~\ref{tab:claude_prompt} for the prompt used in evaluation). 
We perform 10 independent BoN trials with $n=16$ and report the average win rate.
The results are shown in Tab.~\ref{tab:deberta_winrate}. 
The reference PM outperforms CPM-Flan-T5 and standard PM. Interestingly, reference PM shows second best win rate in SHP dataset, highlighting the effectiveness in CPM-GPT-3.5.\fgkII{Dont you just want to say that CPM-GPT-3 outpeforms the reference model in SHP? And why do you say that the reference model outperforms the others if results are mixed?\ptdyII{DY: (27/09/2023) Here IMO CPM-GPT-3.5 outperforms reference model is not a good signal, as it can raise a concern about using reference model as a criteria in Sec~\ref{sec:referece_comparison}}} However, we note that this should be interpreted with caution as CPM-GPT-3.5 is not tested in multiple experiments.

\begin{table}[htb]
\begin{centering}
\begin{tabular}{c|cc}
\toprule
Win Rate & \multicolumn{1}{c}{HH-RLHF} & \multicolumn{1}{c}{SHP}\tabularnewline
\midrule
CPM-GPT-3.5 & 0.810\quad\quad(.) & \textbf{0.672}\quad\quad(.)\tabularnewline
CPM-Flan-T5 & 0.742 (0.034) & 0.580 (0.045)\tabularnewline
Standard PM & 0.588 (0.030) & 0.564 (0.037)\tabularnewline
\midrule
reference PM & \textbf{0.829} (0.050) & 0.619 (0.060)\tabularnewline
\bottomrule
\end{tabular}
\par\end{centering}
\caption{Comparison of win rate between CPM and reference PM externally fitted on multiple datasets. Reference PM shows high win rate in HH-RLHF dataset, and second best win rate in SHP dataset.
\label{tab:deberta_winrate}}
\end{table}

\section{Prompt used}\label{app:prompts}

\begin{table}[htb]
\begin{verbatim}

You will be given a conversation between a human and an AI assistant. 
Your job is to judge whether assistant's reply {attribute_desc}. 
Score that on a scale from 1 to 10 where 1 means {attr_min} and 10 
means {attr_max}. Here's the conversation:

Human:
{history}

Assistant:
{reply}

Answer by outputting a number from 1 to 10 (and nothing else).

Answer:\end{verbatim}

\caption{Prompt template used for  HH-RLHF dataset to rate the response under given attribute.\label{tab:hh-prompt}}

\end{table}

\begin{table}[htb]
\begin{verbatim}

You will be given a Reddit post and a reply. Your job is to judge 
whether the reply {attribute_desc}. Score that on a scale from 1 
to 10 where 1 means {attr_min} and 10 means {attr_max}.

POST:
{query}

Reply:
{reply}

Answer by outputting a number from 1 to 10 (and nothing else).

Answer:\end{verbatim}

\caption{Prompt template used for  SHP dataset to rate the response under given attribute.\label{tab:shp-prompt}}

\end{table}

\begin{table}[htb]
{\footnotesize{}}%
\begin{tabular}{c|c|c}
\toprule
Feature name & Attribute & Description \tabularnewline
\midrule\multirow{3}{*}{{\footnotesize{}helpfulness}} & {\footnotesize{}attribute\_desc} & {\footnotesize{}is helpful for the original poster}\tabularnewline
 
 & {\footnotesize{}attr\_min} & {\footnotesize{}not helpful}\tabularnewline
 
 & {\footnotesize{}attr\_max} & {\footnotesize{}very helpful}\tabularnewline
 
\midrule\multirow{3}{*}{{\footnotesize{}specificity}} & {\footnotesize{}attribute\_desc} & {\footnotesize{}is specific enough}\tabularnewline
 
 & {\footnotesize{}attr\_min} & {\footnotesize{}too vague}\tabularnewline
 
 & {\footnotesize{}attr\_max} & {\footnotesize{}very specific}\tabularnewline
 
\midrule\multirow{3}{*}{{\footnotesize{}intent}} & {\footnotesize{}attribute\_desc} & {\footnotesize{}understands the original poster's intent}\tabularnewline
 
 & {\footnotesize{}attr\_min} & {\footnotesize{}failure of understanding}\tabularnewline
 
 & {\footnotesize{}attr\_max} & {\footnotesize{}perfect understanding}\tabularnewline
 
\midrule\multirow{3}{*}{{\footnotesize{}factuality}} & {\footnotesize{}attribute\_desc} & {\footnotesize{}is factually correct}\tabularnewline
 
 & {\footnotesize{}attr\_min} & {\footnotesize{}egregiously incorrect}\tabularnewline
 
 & {\footnotesize{}attr\_max} & {\footnotesize{}fully correct}\tabularnewline
 
\midrule\multirow{3}{*}{{\footnotesize{}easy-to-understand}} & {\footnotesize{}attribute\_desc} & {\footnotesize{}is easy to understand}\tabularnewline
 
 & {\footnotesize{}attr\_min} & {\footnotesize{}very difficult to understand}\tabularnewline
 
 & {\footnotesize{}attr\_max} & {\footnotesize{}very easy to understand}\tabularnewline
 
\midrule\multirow{3}{*}{{\footnotesize{}relevance}} & {\footnotesize{}attribute\_desc} & {\footnotesize{}is relevant to the original poster's question}\tabularnewline
 
 & {\footnotesize{}attr\_min} & {\footnotesize{}off-topic}\tabularnewline
 
 & {\footnotesize{}attr\_max} & {\footnotesize{}very relevant}\tabularnewline
 
\midrule\multirow{3}{*}{{\footnotesize{}readability}} & {\footnotesize{}attribute\_desc} & {\footnotesize{}is easy to read and not too technical for the original
poster}\tabularnewline
 
 & {\footnotesize{}attr\_min} & {\footnotesize{}very difficult to read}\tabularnewline
 
 & {\footnotesize{}attr\_max} & {\footnotesize{}very easy to read}\tabularnewline
 
\midrule\multirow{3}{*}{{\footnotesize{}enough-detail}} & {\footnotesize{}attribute\_desc} & {\footnotesize{}provides enough detail to be helpful}\tabularnewline
 
 & {\footnotesize{}attr\_min} & {\footnotesize{}too little detail}\tabularnewline
 
 & {\footnotesize{}attr\_max} & {\footnotesize{}very detailed}\tabularnewline
 
\midrule\multirow{3}{*}{{\footnotesize{}biased}} & {\footnotesize{}attribute\_desc} & {\footnotesize{}is biased or one-sided}\tabularnewline
 
 & {\footnotesize{}attr\_min} & {\footnotesize{}very biased}\tabularnewline
 
 & {\footnotesize{}attr\_max} & {\footnotesize{}not biased at all}\tabularnewline
 
\midrule\multirow{3}{*}{{\footnotesize{} \makecell{fail-to-consider-\\individual-preferences}}} & {\footnotesize{}attribute\_desc} & {\footnotesize{}fails to consider the original poster's cultural or
individual preferences}\tabularnewline
 
 & {\footnotesize{}attr\_min} & {\footnotesize{}fails to consider the original poster's preferences}\tabularnewline
 
 & {\footnotesize{}attr\_max} & {\footnotesize{}takes into account the original poster's preferences}\tabularnewline
 
\midrule\multirow{3}{*}{{\footnotesize{}repetetive}} & {\footnotesize{}attribute\_desc} & {\footnotesize{}is repetitive}\tabularnewline
 
 & {\footnotesize{}attr\_min} & {\footnotesize{}very repetitive}\tabularnewline
 
 & {\footnotesize{}attr\_max} & {\footnotesize{}not repetitive}\tabularnewline
 
\midrule\multirow{3}{*}{{\footnotesize{}fail-to-consider-context}} & {\footnotesize{}attribute\_desc} & {\footnotesize{}fails to consider the original poster's context}\tabularnewline
 
 & {\footnotesize{}attr\_min} & {\footnotesize{}fails to consider the original poster's context}\tabularnewline
 
 & {\footnotesize{}attr\_max} & {\footnotesize{}takes into account the original poster's context}\tabularnewline

\midrule\multirow{3}{*}{{\footnotesize{}too-long}} & {\footnotesize{}attribute\_desc} & {\footnotesize{}is too long}\tabularnewline
 
 & {\footnotesize{}attr\_min} & {\footnotesize{}too long}\tabularnewline
 
 & {\footnotesize{}attr\_max} & {\footnotesize{}not too long}\tabularnewline
\bottomrule

\end{tabular}{\footnotesize\par}

\caption{Features and descriptions used in Tab.~\ref{tab:hh-prompt} and \ref{tab:shp-prompt}.\label{tab:attributes}}
\end{table}

\begin{table}[htb]
\begin{verbatim}

Human:
You are a helpful and precise assistant for checking the quality of 
the answer. We would like to request your feedback on the performance 
of two AI assistants in response to the user question displayed below.

[Question]
{query}
[The Start of Assistant 1’s Answer]
{reply1}
[The Start of Assistant 2’s Answer]
{reply2}

Please rate the helpfulness, relevance, accuracy, level of
details of their responses.
Each assistant receives an overall score on a scale of 1 to 10, where
a higher score indicates better overall performance.
Please first output a single line containing only two values indicating
the scores for Assistant 1 and 2, respectively. The two scores are 
separated by a space. In the subsequent line, please provide a
comprehensive explanation of your evaluation, avoiding any potential
bias and ensuring that the order in which the responses were presented 
does not affect your judgment.

Assistant:
\end{verbatim}

\caption{Prompt template to rate the writing quality of the candidate
assistant model.\label{tab:claude_prompt}}

\end{table}

\newpage

\section{Additional model interpretability}

\begin{figure}[htb]
\begin{centering}
\begin{tabular}{cc}
{\includegraphics[width=0.5\columnwidth]{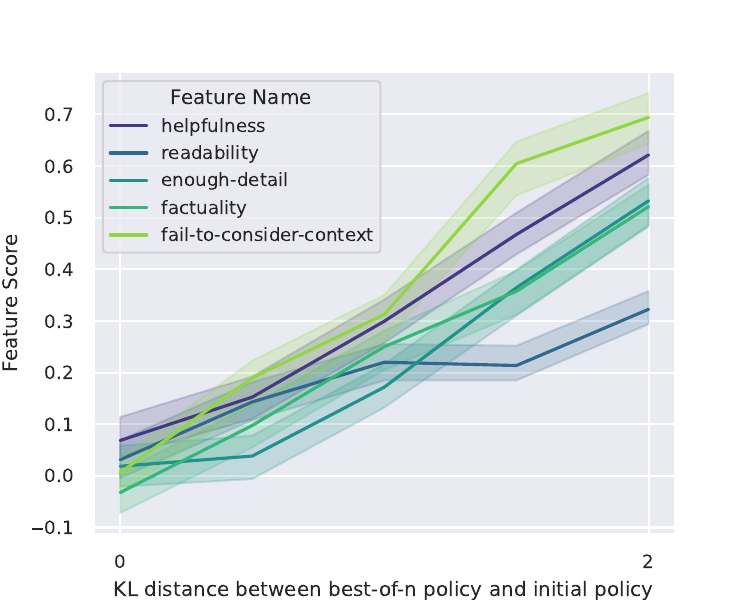}}  & {\includegraphics[width=0.5\columnwidth]{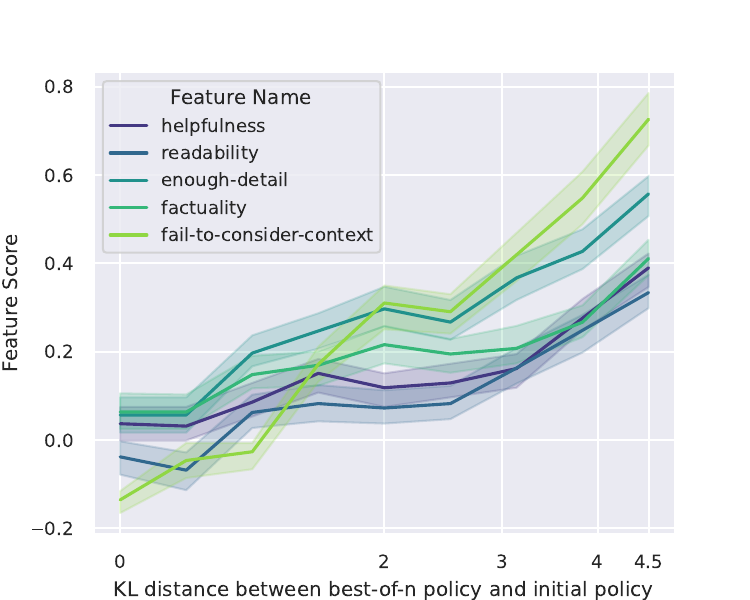}}  \tabularnewline
\end{tabular}
\par\end{centering}
\caption{Feature distribution of BoN experiment (left: CPM-GPT-3.5, right: CPM-Flan-T5). Note that the $x$-ax\ptmdII{e}s are different. \ptmdII{Here the KL distance of the BoN distribution from the initial distribution $a(x)$ is computed as $\text{KL}_\text{BoN}=\log n -\frac{n-1}{n}$ \citep{nakano2021webgpt}.}
\label{fig:feature_distn}}
\end{figure}

\fdyII{(26/09/2023) I've delegated "Feature distribution of policy" to appendix here}
\paragraph{Feature distribution of policy}
Fig.~\ref{fig:feature_distn} shows how the feature values of the responses 
change
\ptdyI{in $\text{BoN}(a,\text{CPM},n)$ distribution with Flan-T5-Large as initial LM $a$,
as a function of KL divergence $\text{KL}_\text{BoN}$. 
This simulates how the feature distribution changes}
as the initial policy is optimized with the current CPM
and allows for inspecting which features drive reward maximization at different stages. \ptdyI{The increase in \texttt{readability} is smaller than other features such as \texttt{enough-detail} and \texttt{fail-to-consider-context}. This means that policy  shifts towards generating responses with more detailed information, rather than focusing on generating readable responses. Additionally, after a certain $n$, the slope of \texttt{fail-to-consider-context} increases, and it achieves the highest score at the end of optimization. }
This means that optimization pressure focuses on not failing to consider the context
and avoiding unexpected lengthy responses. 
\fmdI{(25/09/2023) In the table (and also in the previous table) the name of the feature `fail to consider context' evokes that this is a feature to avoid, but in the table it looks like a positive feature. Should we rename the feature in the tables as "not failing to consider context" ?\ptdyI{DY:(26/09/2023)Would the footnotes that highlight scores are designed to achieve the higher the better be sufficient?}} 
This decomposition makes it easier for a human to understand why the LM generated such responses.\fmdI{(25/09/2023) I agree with TK that these plots are intriguing but the present description is still unclear to me. What are we really trying to show here ?\ptdyI{DY: (26/09/2023) I meant that we can diagnose what happens in LM during alignment tuning. I've tried to make possible intuitive explanation of the feature distribution.}} 

\section{Additional tables and figures}\label{app:additional_figures}

\begin{figure}[ht]
\centering
\begin{tabular}{cc}
(a) HH-RLHF dataset\\
\centerline{\includegraphics[width=\columnwidth]{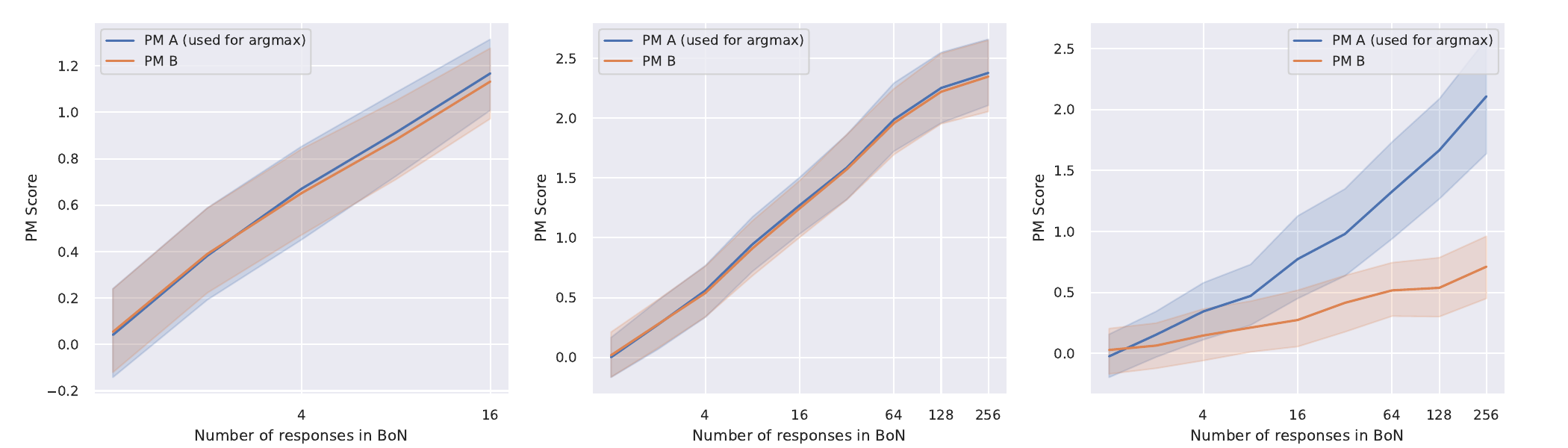}}\\
(b) SHP dataset\\
\centerline{\includegraphics[width=\columnwidth]{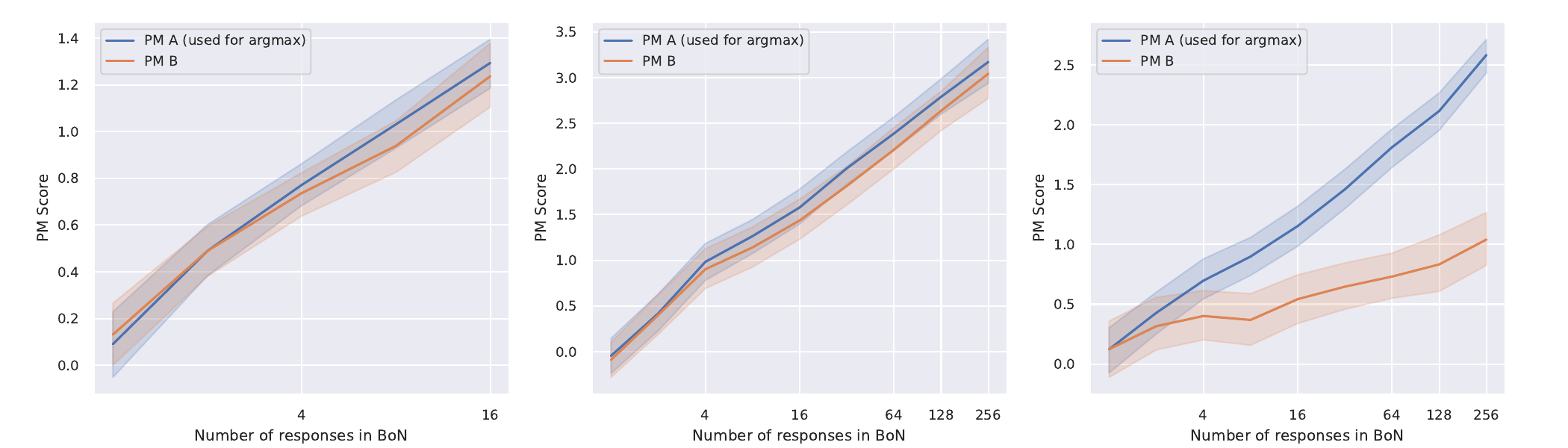}}
\end{tabular}
\caption{
BoN comparison over two models fitted independently in same condition (left: CPM-GPT-3.5, middle: CPM-Flan-T5, right: standard PM) The PM A with blue line indicates the PM used for selection in BoN.}\label{fig:model-variance-app}
\end{figure}

\begin{table}[htb]
\centering{}%
\begin{tabular}{ccc}
\toprule
 & HH-RLHF & SHP\tabularnewline
\midrule
CPM-GPT-3.5 & \textbf{0.997} & \textbf{0.981}\tabularnewline

CPM-Flan-T5 & 0.926 & 0.928\tabularnewline

Standard PM & 0.665 & 0.057\tabularnewline
\bottomrule
\end{tabular}\caption{
\ptdyI{Rank correlation between gold PM scores and proxy PM scores in BoN experiment. For each PM technique used to fit the proxy PM, we calculate and average PM scores over samples from $\text{BoN}(a,\text{PM}_\text{proxy},n)$, and compute the rank correlation between the averaged gold and proxy PM scores over different $n$.}
\label{tab:overoptim_corr}}
\end{table}

\begin{figure}[htb]
\begin{centering}
\begin{tabular}{cc}
{\includegraphics[width=0.5\columnwidth]{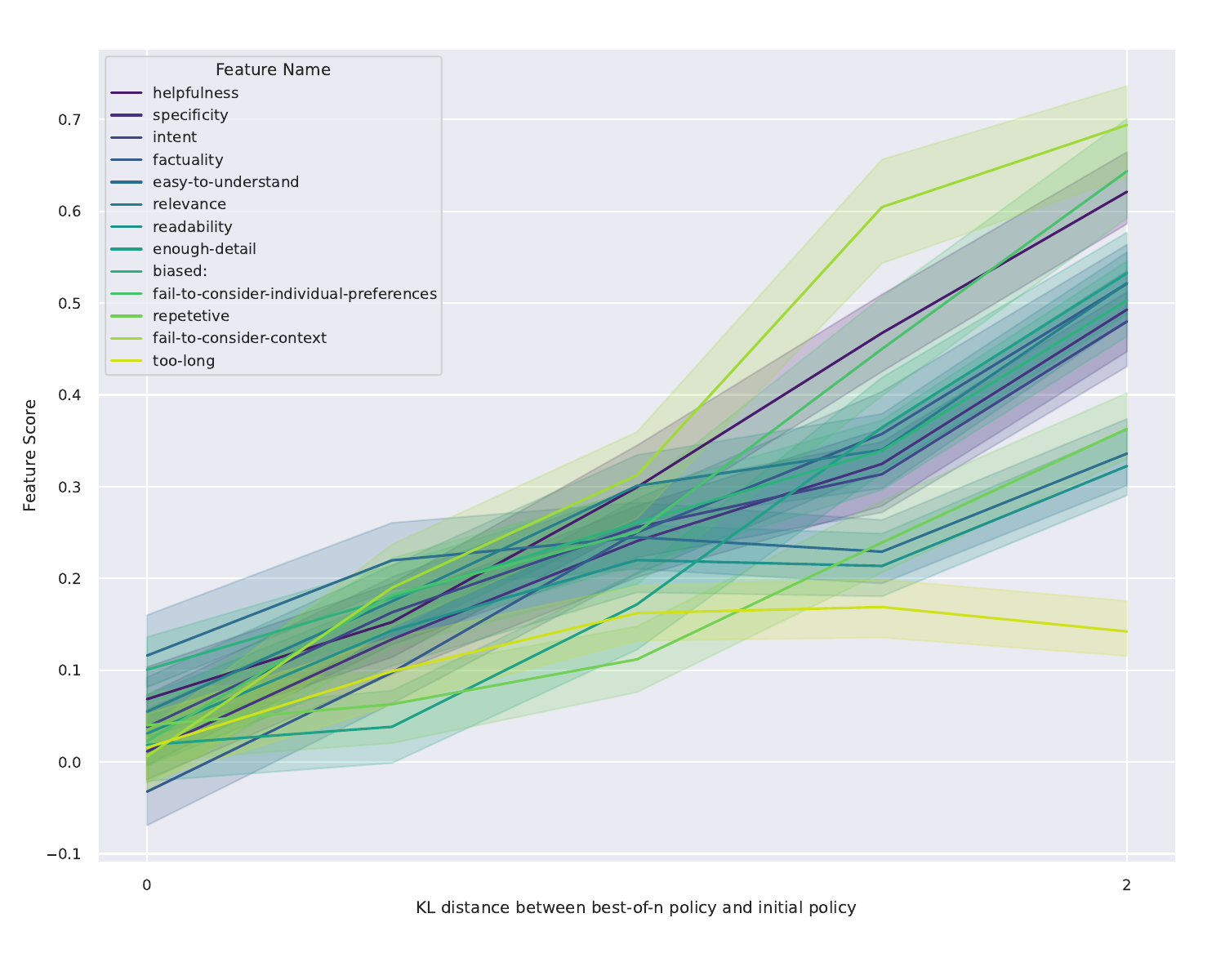}}  & {\includegraphics[width=0.5\columnwidth]{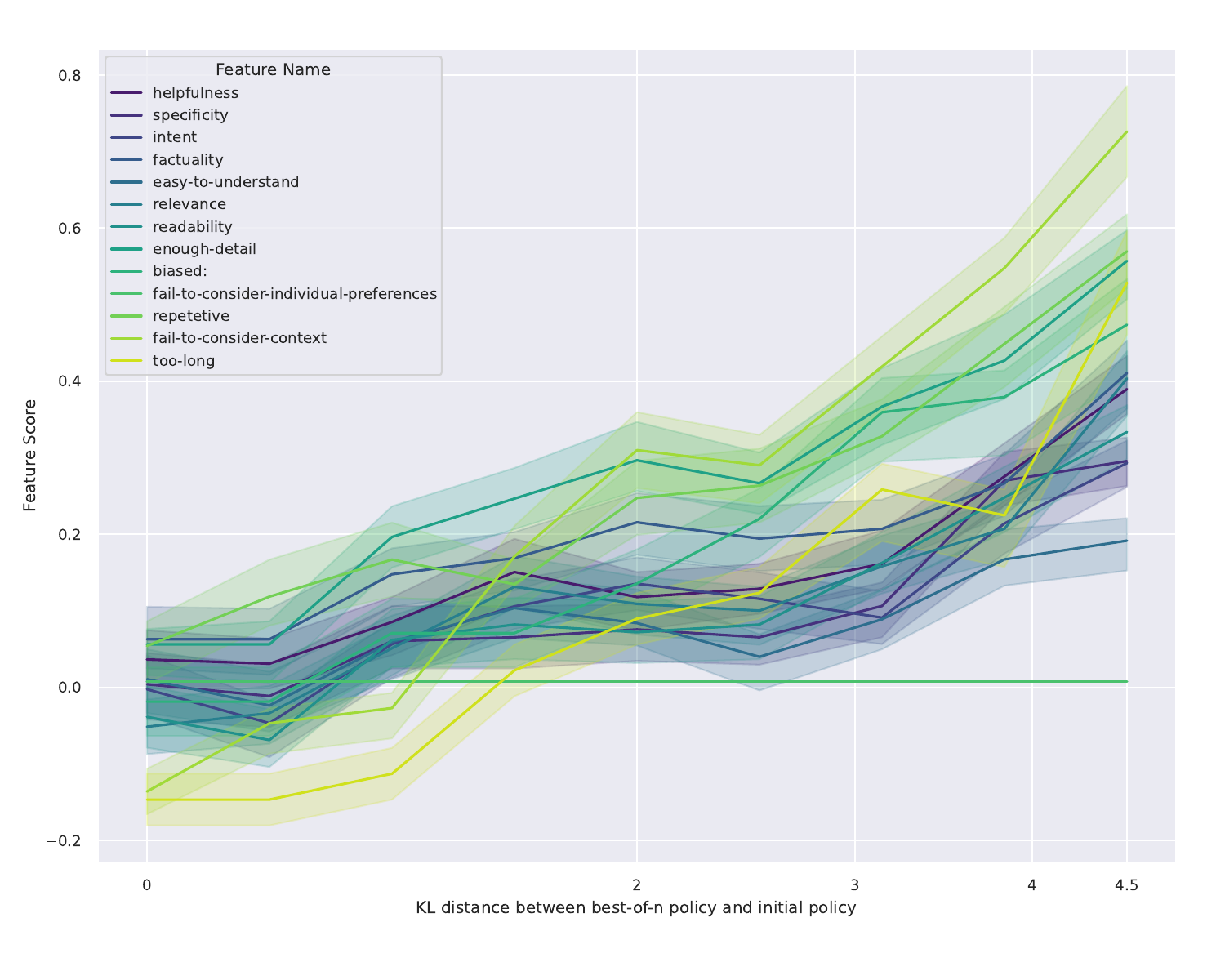}}  \tabularnewline
\end{tabular}
\par\end{centering}
\caption{Feature distribution of BoN experiment (left: CPM-GPT-3.5, right: CPM-Flan-T5). Note that the $x$-ax\ptmdII{e}s are different. \ptmdII{Here the KL distance of the BoN distribution from the initial distribution $a(x)$ is computed as $\text{KL}_\text{BoN}=\log n -\frac{n-1}{n}$ \citep{nakano2021webgpt}.
}}
\end{figure}

\begin{table}[htb]
\begin{centering}
\begin{tabular}{cccc}
\toprule
\multicolumn{2}{c}{CPM-GPT-3.5} & \multicolumn{2}{c}{CPM-Flan-T5}\tabularnewline
\cmidrule(lr){1-2}\cmidrule(lr){3-4}
Feature & Coefficient & Feature & Coefficient\tabularnewline
\midrule
helpfulness & 0.246 & fail-to-consider-context & 0.420\tabularnewline

enough-detail & 0.235 & enough-detail & 0.244\tabularnewline

factuality & 0.187 & factuality & 0.227\tabularnewline

readability & 0.184 & biased & 0.178\tabularnewline

token\_length & 0.101 & easy-to-understand & 0.124\tabularnewline

specificity & 0.094 & specificity & 0.106\tabularnewline

biased & 0.086 & too-long & 0.081\tabularnewline

relevance & 0.071 & token\_length & 0.075\tabularnewline

easy-to-understand & 0.069 & helpfulness & 0.037\tabularnewline

fail-to-consider-context & 0.043 & intent & 0.024\tabularnewline

too-long & 0.016 & repetetive & 0.015\tabularnewline

repetetive & 0.014 & \makecell{fail-to-consider-\\individual-preferences} & -0.042\tabularnewline

intent & -0.008 & relevance & -0.056\tabularnewline

\makecell{fail-to-consider-\\individual-preferences} & -0.056 & readability & -0.120\tabularnewline

\bottomrule
\end{tabular}
\par\end{centering}
\caption{Fitted coefficient of CPM on HH-RLHF dataset. 
\label{tab:full_fitted_coeff}}
\end{table}

\begin{table}[htb]
\centering{}%
\begin{tabular}{lccccc}
\toprule
 & \multicolumn{5}{c}{Features}\tabularnewline
\multicolumn{1}{c}{Dialogue} & {\footnotesize{}\rotatebox[origin=c]{90}{helpfulness}} & {\footnotesize{}\rotatebox[origin=c]{90}{readability}} & {\footnotesize{}\rotatebox[origin=c]{90}{enough-detail}} & {\footnotesize{}\rotatebox[origin=c]{90}{factuality}} & {\footnotesize{}\rotatebox[origin=c]{90}{\makecell{fail-to-consider-\\context}}}\tabularnewline
\midrule
{\footnotesize{}Q: why is ramon laguarta a bad ceo?} \tabularnewline
    
\makecell[l]{\footnotesize{}R1: a bad businessman} & {\footnotesize{}-0.890} & {\footnotesize{}-1.041} & {\footnotesize{}-0.569} & {\footnotesize{}-0.851} & {\footnotesize{}-0.239}\tabularnewline
    
{\footnotesize{}R2: Manager} & {\footnotesize{}-0.890} & {\footnotesize{}-1.682} & {\footnotesize{}-1.063} & {\footnotesize{}-0.851} & {\footnotesize{}-0.710}\tabularnewline
\midrule
{\footnotesize{}Q: What is a blizzard warning?} \tabularnewline

{\makecell[l]{\footnotesize{}R1: Blizzard warnings are issued by the National \\\footnotesize{}Weather Service to warn against snow, ice or strong \\\footnotesize{}winds.}} & {\footnotesize{}1.909} & {\footnotesize{}0.880} & {\footnotesize{}2.395} & {\footnotesize{}1.690} & {\footnotesize{}3.059}\tabularnewline

{\makecell[l]{\footnotesize{}R2: Blizzards can cause extremely dangerous \\\footnotesize{}driving conditions, with heavy snow, ice and sleet.}} & {\footnotesize{}1.559} & {\footnotesize{}0.88} & {\footnotesize{}1.901} & {\footnotesize{}1.408} & {\footnotesize{}2.117}\tabularnewline

\midrule
{\makecell[l]{\footnotesize{}Q: Is climate change raising temperatures above \\\footnotesize{}their normal levels?}} \tabularnewline
    
{\makecell[l]{\footnotesize{}R1: Climate change increases global temperature.}} & {\footnotesize{}1.559} & {\footnotesize{}0.559} & {\footnotesize{}1.901} & {\footnotesize{}1.690} & {\footnotesize{}-0.239}\tabularnewline

{\makecell[l]{\footnotesize{}R2: Earth is expected to experience a warmer climate \\\footnotesize{}by the year 2050, the hottest on record, based on the \\\footnotesize{}International Solar System}} & {\footnotesize{}0.509} & {\footnotesize{}-0.721} & {\footnotesize{}0.419} & {\footnotesize{}0.278} & {\footnotesize{}-0.239}\tabularnewline
\midrule
{\makecell[l]{\footnotesize{}Q: I'd like to evaluate some art perhaps. Can you tell \\\footnotesize{}me some history about Mona Lisa, the painting?}} \tabularnewline
    
{\makecell[l]{\footnotesize{}R1: The Mona Lisa is a 16th century Italian portrait \\\footnotesize{}by Gian Lorenzo Bernini.}} & {\footnotesize{}-0.89} & {\footnotesize{}-1.041} & {\footnotesize{}-0.569} & {\footnotesize{}-0.851} & {\footnotesize{}-0.710}\tabularnewline
    
{\makecell[l]{\footnotesize{}R2: In 1853, the artist Leonardo da Vinci created the \\\footnotesize{}portrait of the Madre della Signore.}} & {\footnotesize{}-0.540} & {\footnotesize{}-1.041} & {\footnotesize{}-0.075} & {\footnotesize{}-0.569} & {\footnotesize{}-0.239}\tabularnewline
\bottomrule
\end{tabular}\caption{Examples of feature values of CPM-GPT-3.5. 
Each feature value can provide information about which aspects of the response are good or bad. \label{tab:feature_values2}}
\end{table}

\clearpage
\reb{\section{Additional results related to questions from reviewers}}

\subsection{\reb{Correlations between features} \label{app:feature_correlation}}

\begin{figure}[ht]
\begin{centering}
\begin{tabular}{cc}
{\includegraphics[width=1\columnwidth]{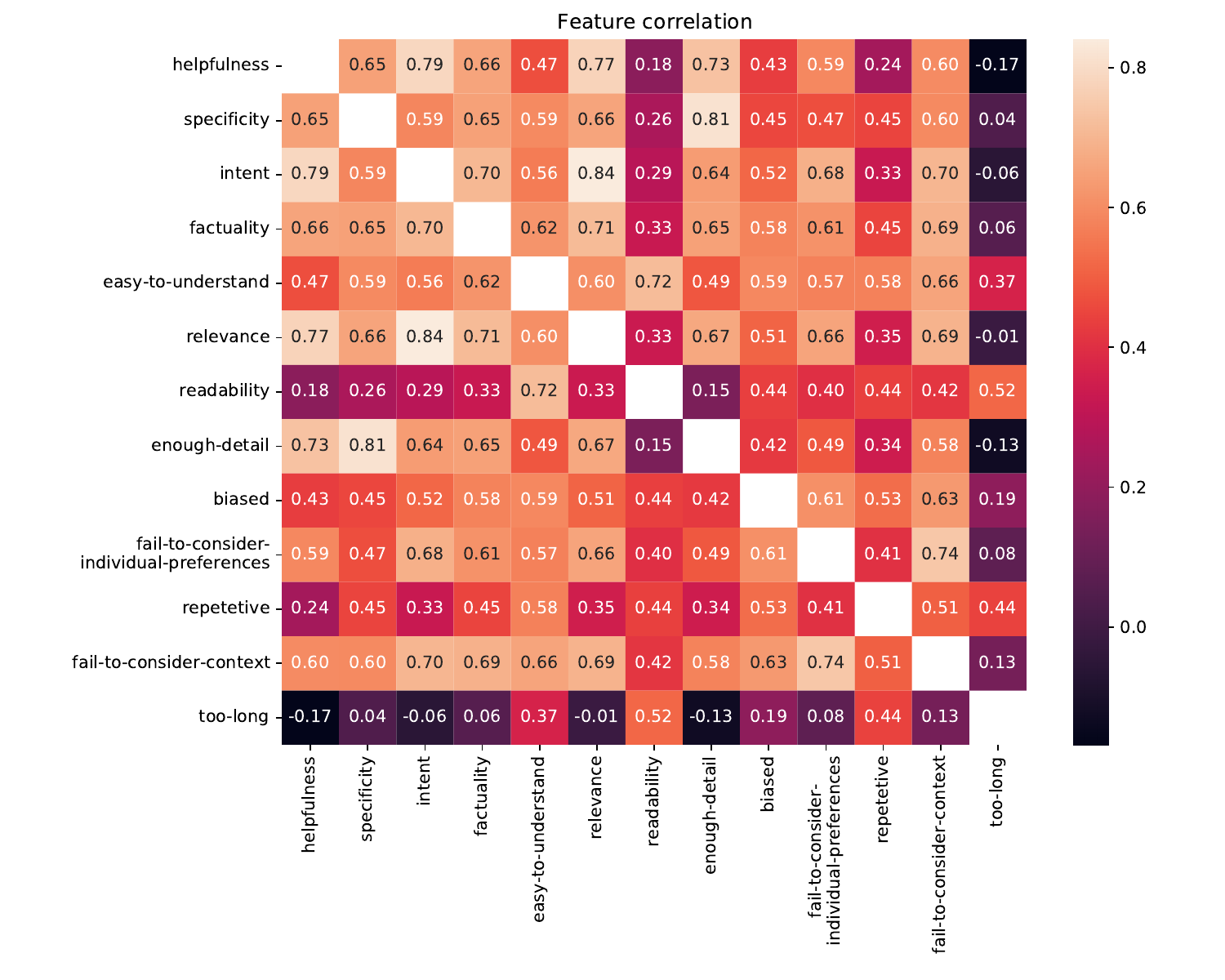}}  \tabularnewline
\end{tabular}
\par\end{centering}
\caption{\reb{Full matrix of feature correlations.
\label{fig:feature-correlation-large}}}
\end{figure}

\reb{To investigate the diversity of features, we compute correlations between features in HH-RLHF dataset. The Figure~\ref{fig:feature-correlation-large} shows that the features have mostly positive correlations, as some of them addressing similar dimensions. }
\fmdIII{(21/11/2023) A minimal text describing this figure would be useful.}

%
%
%
%

\subsection{\reb{Feature scaling trend}\label{app:feature_scaling}}
\reb{To investigate the effect of the number $k$ of features, we gradually increase $k$ and check the win-rate of CPM-Flan-T5 with $k$ features.
For this, we order the features based on their importance in Table~\ref{tab:full_fitted_coeff}, and then assess how the performance of the CPM — measured in terms of ‘win-rate’ quality as in Section~\ref{sec:quality_eval} — varies with $k$ when we keep only the first $k$ most important features. 
Note that regardless of its coefficient rank, we put ‘helpfulness’ first in the ordered list, so that we can compare the case of “prompted PM with one holistic feature” and “compositional PM with $k$ features”.
\\
The ordered feature list is: \texttt{helpfulness, fail-to-consider-context, enough-detail, factuality, length, biased, easy-to-understand, specificity, too-long, intent, repetitive, fail-to-consider-individual-preferences, relevance, readability}. The win-rate averaged for 5 trials is described in Table~\ref{fig:feature-selection}.}

\reb{The table suggests that the single holistic feature ‘helpfulness’ obtains a reasonable win-rate (0.707) on its own,%
\footnote{\reb{One reviewer made the interesting observation that win-rate of the prompted PM with one holistic feature ‘helpfulness’ still comes out ahead that of standard PM (Table~\ref{fig:hh_model_scaling}). 
We hypothesize that the superior performance here of the holistic PM over the standard PM is due to the fact that our preference dataset may not be large enough for the standard PM to achieve robust performance, while the prompted PM utilizes the capabilities of a generic LLM, trained over a huge dataset. }}
but falls short of using the combination of all features (0.742). This suggests that decomposing the features can have additional benefit for capturing the preference. Second, Table~\ref{fig:feature-selection} shows that the performance of CPM with $k=14$ is worse than that of CPM with $k=6$ (0.754). This might be related to the overlap between features. However, the performance gap between $k=14$ and $k=6$ is small, as we employ a regularization term when fitting the logistic classifier.}

\begin{table}[htb]
\centering{}%
\begin{tabular}{cc}
\toprule 
Number of features $k$ & Win Rate \tabularnewline
\midrule
$k=1$ & 0.707 (0.030) \tabularnewline
$k=3$ & 0.715 (0.024) \tabularnewline
$k=6$ & 0.754 (0.038) \tabularnewline
$k=10$ & 0.735 (0.037) \tabularnewline
$k=14$ & 0.742 (0.034) \tabularnewline
\bottomrule
\end{tabular}\caption{
\reb{Win rate of CPM-Flan-T5 over initial generation after BoN sampling based on each PM with different number of features. 
We independently conduct $10$ rounds of BoN($n=16$) samplings and report the average win rate along with standard error.
\label{fig:feature-selection}}}
\end{table}

\subsection{\reb{\ptdyIII{Evaluation with paraphrased} prompts}}\label{app:augmented_prompts}\fdyIII{(21/11/2023)Augmented prompts? \ptmdIII{I think paraphrased prompts is better.}}
\fmdIII{(20/11/2023) I am not sure what this section title refers to...\ptdyIII{DY: (21/11/2023) I've updated the contents. Is it clear to others?} it's great, thanks}

\reb{To further investigate the impact of various prompts and the robustness of the CPM's performance on prompts, we employed GPT-3.5 to paraphrase \ptmdIII{each of} the original descriptions in Table~\ref{tab:attributes}, \ptmdIII{resulting in Table~\ref{tab:augmented_attributes}}.}\fmdIII{(21/11/2023) The caption of this Table should be checked, it seems to be missing a ref to Table 7.\ptdyIII{I've added it} Thanks}

\reb{We evaluated the CPM's performance \ptmdIII{based on this second table,} using the 'win-rate' quality metric described in Section~\ref{sec:quality_eval}. The average win rate of CPM-Flan-T5 across five independent trials was $0.717$ with a standard error of $0.023$, which is not statistically different from the original performance in Table~\ref{tab:claude_eval}, ($0.742$ with a standard error of $0.034$). This indicates that the CPM's performance 
\ptmdIII{shows some robustness relative to}
the specific prompt used.}

\begin{table}[htb]
    {\footnotesize{}}%
    \begin{tabular}{c|c|c}
    \toprule
    Feature name & Attribute & Description \tabularnewline
    \midrule\multirow{3}{*}{{\footnotesize{}helpfulness}} & {\footnotesize{}attribute\_desc} & {\footnotesize{}provides valuable assistance to the original poster}\tabularnewline
     
     & {\footnotesize{}attr\_min} & {\footnotesize{}no assistance}\tabularnewline
     
     & {\footnotesize{}attr\_max} & {\footnotesize{}excellent assistance}\tabularnewline
     
    \midrule\multirow{3}{*}{{\footnotesize{}specificity}} & {\footnotesize{}attribute\_desc} & {\footnotesize{}is detailed and precise}\tabularnewline
     
     & {\footnotesize{}attr\_min} & {\footnotesize{}overly vague}\tabularnewline
     
     & {\footnotesize{}attr\_max} & {\footnotesize{}highly specific}\tabularnewline
     
    \midrule\multirow{3}{*}{{\footnotesize{}intent}} & {\footnotesize{}attribute\_desc} & {\footnotesize{}accurately grasps the original poster's intent}\tabularnewline
     
     & {\footnotesize{}attr\_min} & {\footnotesize{}misinterprets the original poster's intent}\tabularnewline
     
     & {\footnotesize{}attr\_max} & {\footnotesize{}perfectly understands the original poster's intent}\tabularnewline
     
    \midrule\multirow{3}{*}{{\footnotesize{}factuality}} & {\footnotesize{}attribute\_desc} & {\footnotesize{}is based on accurate and verifiable information}\tabularnewline
     
     & {\footnotesize{}attr\_min} & {\footnotesize{}blatantly incorrect}\tabularnewline
     
     & {\footnotesize{}attr\_max} & {\footnotesize{}entirely accurate}\tabularnewline
     
    \midrule\multirow{3}{*}{{\footnotesize{}easy-to-understand}} & {\footnotesize{}attribute\_desc} & {\footnotesize{}is clear and straightforward}\tabularnewline
     
     & {\footnotesize{}attr\_min} & {\footnotesize{}extremely difficult to understand}\tabularnewline
     
     & {\footnotesize{}attr\_max} & {\footnotesize{}exceptionally easy to understand}\tabularnewline
     
    \midrule\multirow{3}{*}{{\footnotesize{}relevance}} & {\footnotesize{}attribute\_desc} & {\footnotesize{}directs addresses the original poster's query}\tabularnewline
     
     & {\footnotesize{}attr\_min} & {\footnotesize{}entirely irrelevant}\tabularnewline
     
     & {\footnotesize{}attr\_max} & {\footnotesize{}highly relevant}\tabularnewline
     
    \midrule\multirow{3}{*}{{\footnotesize{}readability}} & {\footnotesize{}attribute\_desc} & {\footnotesize{}is written in a style appropriate for the original poster's level of understanding}\tabularnewline
     
     & {\footnotesize{}attr\_min} & {\footnotesize{}extremely difficult to read}\tabularnewline
     
     & {\footnotesize{}attr\_max} & {\footnotesize{}exceptionally easy to read}\tabularnewline
     
    \midrule\multirow{3}{*}{{\footnotesize{}enough-detail}} & {\footnotesize{}attribute\_desc} & {\footnotesize{}provides a sufficient level of detail to be helpful}\tabularnewline
     
     & {\footnotesize{}attr\_min} & {\footnotesize{}insufficient detail}\tabularnewline
     
     & {\footnotesize{}attr\_max} & {\footnotesize{}comprehensive level of detail}\tabularnewline
     
    \midrule\multirow{3}{*}{{\footnotesize{}biased}} & {\footnotesize{}attribute\_desc} & {\footnotesize{}presents an objective and impartial perspective}\tabularnewline
     
     & {\footnotesize{}attr\_min} & {\footnotesize{}strong bias or one-sidedness}\tabularnewline
     
     & {\footnotesize{}attr\_max} & {\footnotesize{}completely unbiased}\tabularnewline
     
    \midrule\multirow{3}{*}{{\footnotesize{} \makecell{fail-to-consider-\\individual-preferences}}} & {\footnotesize{}attribute\_desc} & {\footnotesize{}fails to consider the original poster's cultural or
    individual preferences}\tabularnewline
     
     & {\footnotesize{}attr\_min} & {\footnotesize{}fails to consider the original poster's preferences}\tabularnewline
     
     & {\footnotesize{}attr\_max} & {\footnotesize{}carefully considers the original poster's preferences}\tabularnewline
     
    \midrule\multirow{3}{*}{{\footnotesize{}repetetive}} & {\footnotesize{}attribute\_desc} & {\footnotesize{}avoids unnecessary repetition}\tabularnewline
     
     & {\footnotesize{}attr\_min} & {\footnotesize{}excessively repetitive}\tabularnewline
     
     & {\footnotesize{}attr\_max} & {\footnotesize{}not repetitive}\tabularnewline
     
    \midrule\multirow{3}{*}{{\footnotesize{}fail-to-consider-context}} & {\footnotesize{}attribute\_desc} & {\footnotesize{}fails to consider the original poster's situation and background}\tabularnewline
     
     & {\footnotesize{}attr\_min} & {\footnotesize{}fails to consider the original poster's context}\tabularnewline
     
     & {\footnotesize{}attr\_max} & {\footnotesize{}appropriately considers the original poster's context}\tabularnewline
    
    \midrule\multirow{3}{*}{{\footnotesize{}too-long}} & {\footnotesize{}attribute\_desc} & {\footnotesize{}is concise and avoids unnecessary length}\tabularnewline
     
     & {\footnotesize{}attr\_min} & {\footnotesize{}excessively long}\tabularnewline
     
     & {\footnotesize{}attr\_max} & {\footnotesize{}appropriately concise}\tabularnewline
    \bottomrule
    
    \end{tabular}{\footnotesize\par}
    
    \caption{\reb{Paraphrased features augmented from the original descriptions in Table~\ref{tab:attributes}. Those features are used with the template in Table ~\ref{tab:hh-prompt}.\label{tab:augmented_attributes}}}
    \end{table}

\end{document}